\documentclass[twoside]{article}
\usepackage{ecj,palatino,epsfig,latexsym,natbib}

\usepackage{pgfplotstable}
\usepackage{xcolor}
\usepackage{varwidth}
\pgfplotsset{compat=1.14}

\usepackage{algorithm}
\usepackage[noend]{algpseudocode}

\usepackage{amsmath}
\usepackage{subfigure}
\usepackage{graphicx}
\usepackage{multirow}
\usepackage{enumerate}
\usepackage{booktabs}
\usepackage{algpseudocode}
\usepackage{amsmath}

\usepackage[breaklinks,hidelinks]{hyperref}

\algnewcommand{\IIf}[1]{\State\algorithmicif\ #1\ \algorithmicthen}
\algnewcommand{\EndIIf}{\unskip\ \algorithmicend\ \algorithmicif}

\algnewcommand\algorithmicforeach{\textbf{for each}}
\algdef{S}[FOR]{ForEach}[1]{\algorithmicforeach\ #1\ \algorithmicdo}

\begin{document}

\ecjHeader{x}{x}{xxx-xxx}{2019}{A Collaborative Heuristic for UCARP}{J. MacLachlan, Y. Mei, J. Branke, M. Zhang}
\title{\bf Genetic Programming Hyper-Heuristics with Vehicle Collaboration for Uncertain Capacitated Arc Routing Problems}  


\author{\name{\bf Jordan MacLachlan} \hfill \addr{maclacjord@ecs.vuw.ac.nz}\\ 
        \addr{School of Engineering and Computer Science, Victoria University of Wellington, \\
        PO Box 600, Wellington 6140, New Zealand}
\AND
       \name{\bf Yi Mei} \hfill \addr{yi.mei@ecs.vuw.ac.nz}\\
       \addr{School of Engineering and Computer Science, Victoria University of Wellington}
\AND
       \name{\bf Juergen Branke} \hfill \addr{juergen.branke@wbs.ac.uk}\\
        \addr{Warwick Business School, 
        Coventry, CV4 7AL, United Kingdom}
\AND
       \name{\bf Mengjie Zhang} \hfill \addr{mengjie.zhang@ecs.vuw.ac.nz}\\
        \addr{School of Engineering and Computer Science, Victoria University of Wellington}
}

\maketitle

\begin{abstract}

Due to its direct relevance to post-disaster operations, meter reading and civil refuse collection, the Uncertain Capacitated Arc Routing Problem (UCARP) is an important optimisation problem. Stochastic models are critical to study as they more accurately represent the real-world than their deterministic counterparts. Although there have been extensive studies in solving routing problems under uncertainty, very few have considered UCARP, and none consider collaboration between vehicles to handle the negative effects of uncertainty. This paper proposes a novel Solution Construction Procedure (SCP) that generates solutions to UCARP within a collaborative, multi-vehicle framework. It consists of two types of collaborative activities: one when a vehicle unexpectedly expends capacity (\emph{route failure}), and the other during the refill process. Then, we propose a Genetic Programming Hyper-Heuristic (GPHH) algorithm to evolve the routing policy used within the collaborative framework. The experimental studies show that the new heuristic with vehicle collaboration and GP-evolved routing policy significantly outperforms the compared state-of-the-art algorithms on commonly studied test problems. This is shown to be especially true on instances with larger numbers of tasks and vehicles. This clearly shows the advantage of vehicle collaboration in handling the uncertain environment, and the effectiveness of the newly proposed algorithm.

\end{abstract}


\section{Introduction} \label{sec:intro}

The Capacitated Arc Routing Problem (CARP) has been a point of focus in the logistics literature for decades due to its alignment with many real-world problems such as civil refuse collection \citep{Amponsah2004,Lacomme2005,Mei2011b}, winter road gritting \citep{Handa2006,Tagmouti2011} and post-disaster relief \citep{Akbari2017,Celik2015}. Briefly speaking, CARP aims to design a least-cost plan for a fleet of vehicles (each with a limited capacity) to serve a set of edges subject to certain constraints, such as requiring vehicles depart from and return to a depot, and ensuring the total demand served by a vehicle does not exceed its capacity.

Uncertainty is ubiquitous in the real world. For example, in CARP, the demand of an edge (e.g. the amount of waste to be collected on a street) can be uncertain and unknown exactly in advance. The travel cost between two places can also be uncertain, for example depending on real-time traffic conditions. It is very important to take this uncertainty into account to make solutions applicable to real world problems. The Uncertain CARP (UCARP) was proposed \citep{Mei2010} based on this consideration. UCARP includes a wide range of uncertain factors such as the random task demand and travel cost.

The traditional solution optimisation approaches for \emph{static} CARP are not directly applicable to UCARP \citep{Mei2010}, as a solution can fail when a variable's expected and actual values vary. For example, the actual demand of a task can exceed the remaining capacity of the vehicle, and the vehicle cannot fully serve the task as expected.

Routing policies \citep{Weise2012,Liu2017,Mei2018a} are a promising strategy to deal with the uncertain environment in UCARP. In this case, UCARP can be considered an online decision making process. When a vehicle becomes idle, the routing policy is used to give the next instruction to the vehicle based on the latest information. For example, the well known path scanning \citep{Lacomme2004} constructive heuristic can be seen as using a routing policy to generate solutions for UCARP in an online fashion. Routing policies do not require predefined solutions, and thus are very flexible in adapting to environmental changes. Furthermore, they can make decisions in real time, which suits the real-world scenario. For example, the operating centre (with the policy) supplies the vehicle with its next instruction as soon as it becomes idle.

Manually designing effective routing policies is very time consuming and relies heavily on domain expertise. To address this issue, Genetic Programming Hyper-Heuristic (GPHH) has been employed to automatically design heuristics; e.g. dispatching rules in dynamic job shop scheduling \citep{Branke2016,Nguyen2017}, online bin packing rules \citep{Burke2010} and routing policies for uncertain vehicle routing \citep{Weise2012,Liu2017,Jacobsen-Grocott2017,MacLachlan2018}. Typically, GPHH evolves a rule that is used in a problem-specific decision making process. Such a decision making process is often called a \emph{meta-algorithm} \citep{Jakobovic2006,martin2015}. For clarity, we redefine this process as an \emph{SCP}. Given a problem instance and a rule, the SCP provides the context such that a feasible solution can be generated. GPHH uses a set of training instances and the predefined SCP to evolve effective rules, so that given an unseen test instance, the rule is expected to generate a good solution, i.e. the SCP acts as a framework through which routing policies can be evaluated. Its ability to automatically and effectively explore the complex, high-dimensional search space of routing policies lends GPHH well to solving problems such as UCARP.

The design of an effective and suitable SCP is crucial when solving UCARP with GPHH. Existing SCPs are mainly based on construction heuristics that build routes by repetitively inserting a task at the end of a route. The routes may be built sequentially \citep{Weise2012,Liu2017} or in parallel in real time \citep{Mei2018a,MacLachlan2018}. However, existing SCPs have a limitation in that they enforce all tasks be served by a single vehicle, at all costs. For example, when a vehicle fails to complete serving a task (i.e. the actual demand is greater than expected), the failed task has to be served by the same vehicle after refill. A more realistic approach in this case could be to allow multiple vehicles to collaborate in the completion of the failed task.

Intuitively, collaboration between vehicles should be beneficial. Some previous studies have shown that even simple collaboration can lead to an improvement of solution quality. For example, \cite{Ak2007} divided the vehicles into pairs, where the second vehicle appends the first's failed tasks to its route in the event of a failure. A number of studies in deterministic routing have explored split-deliveries which allow tasks to be partitioned between multiple vehicles. As with the paired proposal, many of these methods require optimising the routes for each vehicle in advance, which is difficult to do in a realistic, uncertain environment. To the best of our knowledge, there is no existing study that considers vehicle collaboration in a stochastic environment that handles route construction in real-time.

The overall goal of this paper is to solve UCARP more effectively by using GPHH with vehicle collaboration. Specifically, we aim to
\begin{itemize}
  \item develop a novel Solution Construction Procedure framework that considers vehicle collaboration.
  \item propose a GPHH with vehicle Collaboration (GPHH-C), which evolves a routing policy that works in the heuristic framework with vehicle collaboration. 
  \item demonstrate the effectiveness of GPHH-C on a wide range of UCARP instances, and analyse the obtained routing policies and solutions.
\end{itemize}

The following section describes the UCARP problem in detail, followed by an in-depth review of the existing works in this and related fields. Section \ref{sec:dp} introduces the methods used to enable collaboration, and some techniques used to exploit the new environment. Section \ref{sec:exp} discusses the experimental studies, from settings through results, prior to final further analysis in Section \ref{sec:ana}. 

\section{Background} \label{sec:background}

\subsection{Uncertain Capacitated Arc Routing Problem} \label{sec:UCARP}

A UCARP instance can be represented by a connected graph $G = (V,E)$, where $V$ is the set of vertices and $E$ is the set of edges. The edge set is divided into two subsets $E = E_T \cup E_U$. Each $e \in E_T$ is required to be served (the \emph{tasks}), while each edge in $E_U$ is a non-required edge, i.e. $E_T \cap E_U = \emptyset$. Each task $e \in E_T$ has three features: a non-negative random \emph{demand} $\tilde{d}(e)$, a \emph{serving cost} $\delta_s(e) \geq 0$ and positive random \emph{traversal cost} (cost to traverse without serving) $\tilde{\delta}_t(e)$. Both the random demand and random traversal cost have positive expectations. Each non-required edge $e \in E_U$ has a zero demand $d(e) = 0$, a zero serving cost $\delta_s(e) = 0$ and positive random traversal cost $\tilde{\delta}_t(e)$. The \emph{depot} is denoted as $v_0 \in V$. The tasks are to be served by a set of vehicles, each with a finite max capacity $Q$, and remaining $q(k)$ capacity. 

A \emph{sample} of a UCARP instance is obtained by sampling a value for each random variable of the corresponding UCARP instance. For example, a sample $I_\xi$ of the UCARP instance $I$ is obtained by sampling each random demand $d_\xi(e), \forall e \in E_T$ and each random traversal cost $\delta_{t,\xi}(e), \forall e \in E$ under the environment (e.g. random seed) $\xi$. 

A sample of a UCARP instance is similar to a commonly considered \emph{static CARP} instance, but more complex. Unlike its static precursor, in a UCARP instance sample, both the task demand and edge traversal cost are unknown in advance, and are only revealed upon having attempted the serving or traversal of the corresponding edge. Prior to this point, only historical distributions can be used to approximate the values. 

Due to the above online realisation process of the random variables, the following two failures may occur in a solution to a UCARP instance sample.

\begin{itemize}
	\item \emph{Route failure}: the actual demand of the task to be served exceeds the remaining capacity of the vehicle.
	\item \emph{Edge failure}: the edge ahead of the route is inaccessible.
\end{itemize} 

In the case of route failure, the solution needs to be repaired. A typical recourse operator uses the same vehicle to finish the failed task. When route failure occurs, the vehicle returns to the depot to refill its capacity, and then comes back to finish the remaining service of the failed task. In the case of edge failure, one can find a detour (e.g. by Dijkstra's algorithm under the current known environment). In comparison to edge failure, route failure is more challenging and has a greater impact on solution quality. Therefore, in this paper, we focus on tackling the uncertain demand, and thus prioritise route failure. 

A solution to a UCARP instance sample is represented as $S=(X,Y)$. $X=\{X^{(1)}, \dots, X^{(m)}\}$ is a set of vertex sequences, where $X^{(k)}=(x^{(k)}_1, \dots, x^{(k)}_{L_k})$ stands for the $k^{th}$ route and $L_k$ is the number of vertices in route $X^{(k)}$. $Y=\{Y^{(1)}, \dots, Y^{(m)}\}$ is a set of continuous vectors, where $Y^{(k)}=(y^{(k)}_1, \dots, y^{(k)}_{L_k-1})$ ($y^{(k)}_i \in [0, 1]$) is the fraction of demand served at each edge implicitly defined by route $X^{(k)}$. For example, if $y^{(1)}_3= 0.7$, then $(x^{(1)}_3,x^{(1)}_{4})$ is a task and $70\%$ of its demand is served at position 3 of the route $X^{(1)}$. Practical examples of this are given in Tables \ref{tab:GPHH_taskSeqGDB1} \& \ref{tab:GP-PRX_taskSeqGDB1}.

With the above notation, the problem can be formulated as follows.
\begin{align}
\min \;\; & \mathrm{E}_{\xi \in \Xi}[C(S_{\xi})], \label{eq:obj} \\
s.t. \;\; & S_{\xi}[x^{(k)}_1]=S_{\xi}[x^{(k)}_{L_k}]=v_0,\ \forall\ k=1,2,...,m \label{eq:depot} \\
& \sum_{k=1}^{m} \sum_{i=1}^{L_k-1} S_{\xi}[y^{(k)}_i]\cdot S_{\xi}[z^{(k)}_i(e)] = 1,\ \forall\ d_{\xi}(e) > 0, \label{eq:once1} \\
& \sum_{k=1}^{m} \sum_{i=1}^{L_k-1} S_{\xi}[y^{(k)}_i]\cdot S_{\xi}[z^{(k)}_i(e)] = 0,\ \forall\ d_{\xi}(e) = 0, \label{eq:once2} \\
& \sum_{i=1}^{L_k-1}d_{\xi}\left(S_{\xi}, k, i\right) \cdot S_{\xi}[y^{(k)}_i] \leq Q,\ \forall\ k=1, \dots, m, \label{eq:capacity} \\
& \left(S_{\xi}[x^{(k)}_i], S_{\xi}[x^{(k)}_{i+1}]\right) \in E, \label{eq:xdomain} \\
& S_{\xi}[y^{(k)}_i] \in [0, 1], \label{eq:ydomain}
\end{align} where $S_{\xi}[x^{(k)}_i]$ and $S_{\xi}[y^{(k)}_i]$ stand for the $x^{(k)}_i$ and $y^{(k)}_i$ elements in $S_{\xi}$. $S_{\xi}[z^{(k)}_i(e)]$ equals $1$ if $\left(S_{\xi}[x^{(k)}_i], S_{\xi}[x^{(k)}_{i+1}]\right) = e$, and $0$ otherwise. i.e. if the next edge in the route is $e$, return 1. $d_{\xi}\left(S_{\xi}, k, i\right)$ is the actual demand of the edge $\left(S_{\xi}[x^{(k)}_i], S_{\xi}[x^{(k)}_{i+1}]\right)$. 

Eq. (\ref{eq:obj}) is the objective function, which is to minimise the expected total cost $C(S_{\xi})$ of the solution $S_{\xi}$ over all the possible samples $\xi \in \Xi$. Here, $C(S_{\xi})$ is calculated as follows.

\begin{align}
C(S_{\xi}) = \sum_{k=1}^{m}\sum_{i=1}^{L_k-1}\delta_{t,\xi}(S_{\xi}[x^{(k)}_{i}], S_{\xi}[x^{(k)}_{i+1}]) + \sum_{e \in E_T}(\delta_{d}(e) - \delta_{t,\xi}(e)).
\end{align}

Eq. (\ref{eq:depot}) indicates that in all $S_{\xi}$, the routes start and end at the depot. Eqs. (\ref{eq:once1}) and (\ref{eq:once2}) require that each task is served exactly once (the sum total demand fraction served by all collaborating vehicles is 1), while each non-required edge is not served. 
Eq. (\ref{eq:capacity}) is the capacity constraint, and Eqs. (\ref{eq:xdomain}) and (\ref{eq:ydomain}) are the domain constraints.

Note that $S_{\xi}$ varies from one sample to another. For any sample $\xi$, a feasible solution $S_{\xi}$ can be generated by a pre-optimised (robust) solution plus a recourse operator, or a routing rule that gradually builds the solution in an online fashion. 

Note that the commonly considered static CARP is a special case of UCARP, where the variables are deterministic. The significant extra challenge of UCARP over static CARP is that the route failures caused by random task demand can lead to large extra refill cost under traditional recourse policies.

\subsection{Related Work} \label{sec:relatedWork}

The challenges of UCARP come from two aspects. First, the static CARP itself is NP-hard \citep{Belenguer1998} and thus challenging to solve. Second, it is difficult to make proper decisions in the uncertain environment. In the following, we will first introduce the related work from the above two aspects, and then review the existing works for routing, specifically Vehicle Routing Problems (VRPs) and CARP, under uncertainty.

\subsubsection{Approaches for CARP}

The early studies for solving CARP focused on exact approaches, such as integer programming \citep{Belenguer1998} and cutting plane algorithm \citep{Belenguer2003}. The exact approaches can guarantee the optimality of the obtained solutions. However, due to the NP-hardness of CARP, the exact approaches are restricted to small static instances. When the problem size becomes large, these approaches become too time consuming, and not practically applicable.

Contrary to the exact approaches, the constructive heuristic approaches generate approximated solutions from scratch. These approaches cannot guarantee optimality. However, they are much cheaper than the exact approaches, and can often give a reasonably good solution in a very short time. Examples of commonly used construction heuristics for CARP include the path scanning heuristic \citep{Lacomme2004}, augment-merge heuristic \citep{Lacomme2004}, and Ulusoy's split heuristic \citep{Lacomme2004}.

The meta-heuristics are becoming more and more commonly used in recent years. A typical meta-heuristic algorithm starts from one or more solutions, and iteratively improves them (e.g. by crossover, mutation or other local search operators). In this way, meta-heuristic algorithms can usually obtain promising solutions in a given time budget. A meta-heuristic is usually no worse than a constructive heuristic (e.g. it can take the solution of a construction heuristic as an initial solution), and is much more efficient than exact approaches. There have been a number of meta-heuristic algorithms proposed for CARP, including evolutionary algorithms \citep{Lacomme2004,Tang2009,Mei2011,Mei2014,Feng2015}, ant colony optimisation \citep{santos2010improved,xing2011hybrid}, tabu search \citep{Hertz2000,brandao2008deterministic,Mei2009}, guided local search \citep{beullens2003guided} and variable neighbourhood search \citep{polacek2008variable}. The keys to success of these search-based algorithms are the development of the solution representation and search operators to achieve a better trade-off between exploration (diversity) and exploitation (convergence).

\subsubsection{Handling Uncertain Environments}

The existing approaches for handling uncertain or dynamic environments can be mainly divided into three categories \citep{Ouelhadj2009,Nguyen2016CEC}: (1) robust proactive approaches; (2) completely reactive approaches; and (3) predictive-reactive approaches.

The robust proactive approaches typically contain two stages. In the first stage, one or more robust solutions are obtained by optimisation algorithms based on the prediction of the environment. Then, in the second stage, the robust solutions are executed, during which recourse actions are taken to deal with possible failures (e.g. when the capacity expires in the middle of serving a task in UCARP). The success of these approaches heavily relies on accurate prediction of the stochastic environment, and the design of proper recourse operators. Examples of the robust proactive approaches for stochastic vehicle routing include the two-stage stochastic programming with recourse \citep{Kall1994,Christiansen2009,Christiansen2007,Gendreau2016} and the genetic algorithms that optimise robustness as the fitness function \citep{Fleury2004,Fleury2005,Wang2013,Wang2016}.

In contrast with the robust proactive approaches, the completely reactive approaches do not maintain any pre-optimised solution. Instead, they treat the problem as an online decision making process, and build a complete solution step by step using a decision-making rule (e.g. routing policy in UCARP) in a decision making process (i.e. \emph{meta-algorithm}). Some constructive heuristics such as the path scanning heuristic \citep{Lacomme2004} can also be seen as a completely reactive approach.

When developing completely reactive approaches, the key issues are the development of the decision making process (i.e. meta-algorithm) and the rule. The meta-algorithm is problem specific, and is generally designed by human experts. For example, the Johnson's algorithm \citep{johnson1954optimal} can be used as a meta-algorithm for two-machine static job shop scheduling \citep{hunt2014evolving}, and the path scanning heuristic process can be used as a meta-algorithm for UCARP \citep{Weise2012,Liu2017}. The rollout heuristics \citep{AbdAllah2017,Ulmer2018} and Monte Carlo tree search \citep{sabar2015population} can also be used as meta-algorithms. The rule, on the other hand, can be designed either manually or automatically. Early studies focused on manually designing the rules, such as the dispatching rules for job shop scheduling (e.g. \citep{blackstone1982state,holthaus1997efficient}). Recently, automatically designing rules using Genetic Programming Hyper-Heuristics (GPHH) \citep{burke2009exploring} has become a dominant approach to automatically evolving the rules, such as evolving the dispatching rules for job shop scheduling \citep{Branke2016,Nguyen2017}, vehicle routing \citep{Jacobsen-Grocott2017}, and online bin packing \citep{burke2006evolving}. A commonly used GPHH approach evolves a priority function (as a Lisp tree), which is used to select the next candidate task from the current pool (e.g. all the jobs waiting in a machine's queue in job shops scheduling).

The predictive-reactive approaches combine the characteristics of both the robust proactive approaches and the completely reactive approaches. In general, these approaches first obtain a predictive baseline solution (e.g. by proactive approaches), and then reoptimise the solution after a certain time period (e.g. \citep{Montemanni2002,Hanshar2007}) or upon the occurrence of real-time events (e.g. \citep{Chryssolouris2001}). These approaches generally consider both the quality of the predictive baseline solution (\emph{efficacy}) and the degree of change to be made on the baseline solution to adapt to the new environment (\emph{stability}). Most works consider optimising the efficacy and stability in a multi-objective optimisation framework \citep{Leon1994,Fattahi2010,Shen2015}.

\subsubsection{Existing Works for Routing Under Uncertainty}

There have been limited studies considering uncertainty in CARP. \cite{Fleury2004,Fleury2005} proposed a robust proactive evolutionary algorithm for optimising solutions for CARP with stochastic demand. \cite{Christiansen2009} formulated the CARP with stochastic demand as a two-stage stochastic program and developed a branch-and-price algorithm to solve it. The state-of-the-art approach for UCARP is the EDA with stochastic local search (EDASLS) proposed by \cite{Wang2016}, which evolves robust solutions for UCARP. For completely reactive approaches, \cite{Weise2012} proposed a GPHH algorithm to evolve the priority function used in the path scanning heuristic to make routing decisions in real time for CARP with stochastic demand. \cite{Liu2017} improved the GPHH approach by designing a better meta-algorithm and more informative terminal and function sets. \cite{MacLachlan2018} further improved the GPHH algorithm by designing a new problem-specific terminal, which can alleviate the negative effect of route failure on the solution quality.

As the node routing counterpart of CARP, the dynamic and stochastic VRP have been studied much more extensively. Most existing methods for solving VRP and CARP under uncertainty belong to the robust proactive (i.e. optimising a robust solution that, with simple recourse action, shows relatively high quality over all possible environments) \citep{Christiansen2007} or reactive (i.e. formulating the problem as a decision making process, and optimising the decision making policy) \citep{Secomandi2001,Secomandi2003,Bertazzi2018,Ulmer2018,Ulmer2018b,Goodson2016}) categories. More details can be found in the comprehensive surveys \citep{Pillac2013,Oyola2016,Oyola2017,Gendreau2016,Ritzinger2016}.

\subsubsection{Collaborative Routing}

Most recourse actions in past studies have been confined to single vehicle representations. In many real-world cases, however, it is not necessary for recourse to be limited in this way, and it is plausible that vehicles could collaborate to complete tasks. \cite{Ak2007} divide vehicles into pairs (Type I and Type II), and optimise the routes for each using tabu search. Then, when a route failure occurs, the failed customers of Type I routes are appended to the end of the paired Type II route. Several more advanced pairing strategies were proposed by \cite{Lei2012,Erera2009}. A better collaborative representative, however, is that of Split Deliveries \citep{Dror1989,Dror1990,Mullaseril1997} (for reviews, see: \citep{Gulczynski2008,Archetti2012}). In this case, individual tasks may be served by more than a single vehicle, allowing distance-cost savings of up to 50\% on static instances (in extreme cases), provided the Triangle Inequality holds.

Routing with split deliveries has been solved using heuristics with local search \citep{Labadi2008}, meta-heuristics \citep{Belenguer2010}, particle swarm optimisation \citep{Shi2018} and exact methods \citep{Ozbaygin2018}. However, all of these methods operate within a deterministic framework. It is logically foreseeable that the static solution methods utilised to date would face the same transference and performance issues of other static routing methods when applied to an uncertain problem space. We are not aware of any works that consider split deliveries within a stochastic environment.

In summary, many methods have been utilised in solving the array of NP-hard routing problems. In the field of static routing, meta-heuristics are often implemented as a cost-effective technique of generating solutions over their mathematical counterparts which become intractable on large instances. However, static problem representations often fail to accurately portray problems encountered in the real-world, so a number of stochastic alternatives have been suggested. This new uncertainty has required the development of different algorithms, as many of the static methods become ineffective. Unfortunately, despite its direct relevance to the real world, only a few methods have been proposed to solve UCARP to date. EDASLS \citep{Wang2016}, the existing state-of-the-art algorithm, and GPHH \citep{Liu2017}, its main rival, are those directly compared in this work. Finally, existing stochastic routing work has failed to take into account the changes in modern communications technology, or the practicalities of many real-world problems by failing to facilitate vehicle collaboration. This paper attempts to solve this issue by implementing a GPHH for UCARP enabling such collaboration.

\subsection{A Genetic Programming Hyper Heuristic}

Algorithm \ref{alg:GPHH_fw} shows a general framework of GPHH for UCARP \citep{Liu2017, Mei2018a}.

\begin{algorithm}[!ht]
	\caption{The general framework of a GPHH for UCARP}
	\label{alg:GPHH_fw}
	\begin{algorithmic}[1]
    	\State Randomly initialise a population of routing policies;
    	\While{Stopping criteria not met}
	        \State \begin{varwidth}[t]{\linewidth} 
	        Evaluate the routing policies in the current population using an \par \hskip\algorithmicindent appropriate SCP;
	        \end{varwidth}
	        \State \begin{varwidth}[t]{\linewidth}
	        Generate an offspring population by applying \par
	        \hskip\algorithmicindent crossover/mutation/reproduction operators;
	        \end{varwidth}
	        \State \begin{varwidth}[t]{\linewidth}
	        Select the routing policies from the current and offspring populations \par
	        \hskip\algorithmicindent to form the new population;
	        \end{varwidth}
    	\EndWhile
    \end{algorithmic}
\end{algorithm}

In Algorithm \ref{alg:GPHH_fw}, individuals take the form of routing policies (priority trees). At each decision point, a routing policy determines each vehicle's next task from the unserved task set. To evaluate each routing policy, an SCP and a set of training instances are needed. The routing policy is applied via the SCP to generate a solution for each training instance. The fitness of the policy is defined as the average solution quality over the training instances. As a result, the effective design of SCPs is critical to the success of the GPHH process.

Making SCPs more effective is therefore a constant focus of those involved in GPHH literature. In routing problems such as UCARP, phenomena present in the real-world (such as vehicle collaboration) have been previously overlooked that could lead to improvements in performance. The following section describes the chosen method of implementing vehicle collaboration through a novel SCP. 


\section{A New Solution Construction Procedure with Vehicle Collaboration} \label{sec:dp}

Given a UCARP instance sample, a number of vehicles, and a routing policy, the SCP with vehicle collaboration generates a solution to the UCARP instance sample. 


Algorithm \ref{alg:colldp} describes the proposed SCP with vehicle collaboration, which is an event-driven decision making process. Initially, all tasks are unserved and unassigned, i.e. the unserved $\Omega_{\text{unser}}$ and unassigned $\Omega_{\text{unass}}$ task sets are both set to $E_T$, and the remaining demand fraction $\theta (t)$ for each task $t \in E_T$ is set to $1$. All vehicles begin at the depot, i.e. $X^{(k)} = (v_0)$ and $Y^{(k)} = (\:)$, and are empty loaded, i.e. remaining capacity $q(k) = Q$. The initial event queue $\Gamma$ consists of a set of {refill events}, one for each vehicle. Then, decisions for the vehicles are made by triggering the event in $\Gamma$ with earliest start time, updating the state (e.g. $(X, Y)$, $\Omega_{\text{unser}}$, $\Omega_{\text{unass}}$ and $\Gamma$), until $\Gamma$ becomes empty. Finally, the routes of the vehicles are returned.

\begin{algorithm}[!ht]
	\caption{The Solution Construction Procedure with Vehicle Collaboration.}
	\label{alg:colldp}
	\begin{algorithmic}[1]
    	\Require A UCARP instance sample $I_{\xi}$, number of vehicles $m$, a routing policy $h(.)$.
		\Ensure A solution $S_{\xi} = (X_{\xi}, Y_{\xi})$.
        \State $\Omega_{\text{unser}} = E_T, \Omega_{\text{unass}} = E_T$;
        \ForEach {task $t \in E_T$} 
        	$\theta (t) = 1$;
        \EndFor
        \For{$k = 1 \rightarrow m$}
        	$X_{\xi}^{(k)} = (v_0), Y_{\xi}^{(k)} = (\:), q(k) = Q$;   
        \EndFor
        \State Initiliase an empty event queue $\Gamma$;
        \For{$k = 1 \rightarrow m$}
        	add into $\Gamma$ a refill event for vehicle $k$, occurring at time $0$;
        \EndFor
        \While{$\Gamma$ is not empty}
        	\State select the next event $\epsilon \in \Gamma$, and remove it from $\Gamma$;
        	\State trigger $\epsilon$ to update $X_{\xi}$, $Y_{\xi}$, $\Omega_{\text{unser}}$, $\Omega_{\text{unass}}$, $\theta(\cdot)$, $q(\cdot)$, $\Gamma$;
        \EndWhile
        \State \Return $S_{\xi} = (X_{\xi}, Y_{\xi})$;
    \end{algorithmic}
\end{algorithm}

The new SCP has the following three distinct advantages over the SCPs proposed by \cite{Liu2017} and \cite{Mei2018a}: 1) the manner in which the environment is updated in the event of a route failure, via a new collaborative \emph{Serving event}; 2) the recourse process a vehicle undertakes in the event of a route failure, via a new collaborative \emph{Refill event}; 3) the method of estimating the remaining demand of previously failed tasks.

Note that the new SCP can use routing policies designed using any method (e.g. manually or automatically) and any decision making events (collaborative, or otherwise).

There are two types of events during the decision making process, as follows:
\begin{itemize}
    \item \emph{Refill event}: a vehicle returns to the depot to refill its capacity. It has the following two parameters: (1) the event time and (2) the vehicle.
    \item \emph{Serving event}: a vehicle goes to a target task and serves it. It has the following three parameters: (1) the event time, (2) the vehicle and (3) the task to be served.
\end{itemize}

The collaboration between vehicles is implemented in both the refill and serving events. Details of these two events is given in Sub-section \ref{sec:events}.

\subsection{Collaborative Events}
\label{sec:events}

Algorithm \ref{alg:refill} shows the operator to \emph{trigger} a refill event. Per \cite{Liu2017, Mei2018a}, if the vehicle reaches the depot, then its remaining capacity is reset to $Q$, and the policy $h(\cdot)$ is used to decide the next task to serve (lines \ref{line:refill1}--\ref{line:refill2}). Otherwise, the vehicle goes to the next node on the shortest path to the depot (lines \ref{line:refillgo1}--\ref{line:refillgo2}). The novel collaborative component is in lines \ref{line:refillcol1}--\ref{line:refillcol2}, where the vehicle checks every edge on its way to the depot. If an edge is a task that has yet to be completed, then the vehicle serves the task (fully or partially). If the vehicle fully serves a task that has already been assigned to another vehicle, then the refill operator reassigns a new task to said vehicle (lines \ref{line:refillreassign1}--\ref{line:refillreassign2}). There are three scenarios under which this operator is useful: $(1)$ the one provided above, where the remainder of the originally assigned vehicle's trip to the task is saved; $(2)$ where the vehicle can fully serve an unassigned task and effectively invoke zero future traversal cost; and $(3)$ where the partial serving of a task results in further exploration of the graph, important in Section \ref{sec:dem-est}. 

\begin{algorithm}[!ht]
	\caption{Triggering a refill event.}
	\label{alg:refill}
	\begin{algorithmic}[1]
    	\State get the vehicle $k$ and its current location $v_{\text{curr}}$ in $\epsilon$;
        \If{$v_{\text{curr}} = v_0$} \label{line:refill1}
            \State $q(k) = Q$;
            \State identify the candidate tasks $\bar{\Omega} \subseteq \Omega_{\text{unass}}$; \label{line:cand1}
            \If{$\bar{\Omega} = \emptyset$} \Return;
            \EndIf
        	\State select the next task $t^* = \arg\min_{t \in \bar{\Omega}} h(t)$; \label{line:refillrule}
            \State $\Gamma = \Gamma \cup$\{serving event for vehicle $k$ to serve $t^*$\}; \label{line:refill2}
        \Else
        	\State find the next node $v'$ on the way to $v_0$; \label{line:refillgo1}
        	\State $X_{\xi}^{(k)} = (X_{\xi}^{(k)}, v')$;
        	\If{$\theta(v_{\text{curr}}, v') > 0$} \label{line:refillcol1}
        	    \If{$q(k) \geq \theta(v_{\text{curr}}, v') \cdot d_{\xi}(v_{\text{curr}}, v')$}
        	        \State $Y_{\xi}^{(k)} = (Y_{\xi}^{(k)}, \theta(v_{\text{curr}}, v'))$;
        	        \State $q(k) = q(k) - \theta(v_{\text{curr}}, v') \cdot d_{\xi}(v_{\text{curr}}, v')$;
        	        \State $\theta(v_{\text{curr}}, v') = 0$;
        	        \State $\Omega_{\text{unser}} = \Omega_{\text{unser}} \setminus (v_{\text{curr}}, v')$;
        	        \If{$(v_{\text{curr}}, v')$ is assigned to vehicle $k'$} \label{line:refillreassign1}
        	            \State reassign a new task to vehicle $k'$ by $h(\cdot)$;
        	        \EndIf \label{line:refillreassign2}
        	    \Else
        	        \State $\theta' = q(k) / d_{\xi}(v_{\text{curr}}, v')$;
        	        \State $Y_{\xi}^{(k)} = (Y_{\xi}^{(k)}, \theta')$;
        	        \State $q(k) = 0$;
        	        \State $\theta(v_{\text{curr}}, v') = \theta(v_{\text{curr}}, v') - \theta'$;
        	    \EndIf \label{line:refillcol2}
    	    \Else
    	        \State $Y_{\xi}^{(k)} = (Y_{\xi}^{(k)}, 0)$;
        	\EndIf
        	\State $\Gamma = \Gamma \cup$\{refill event for vehicle $k$\}; \label{line:refillgo2}
        \EndIf
    \end{algorithmic}
\end{algorithm}

Algorithm \ref{alg:serving} shows the operator to trigger a serving event. The vehicle repeatedly steps to the updated next node until it arrives the head node of the target task (lines \ref{line:servego1}--\ref{line:servego2}). Note that if the vehicle passes the depot on the way to the target task, its capacity is refilled (line \ref{line:servingrefill}). The serving starts when the vehicle arrives the head node of the target task (line \ref{line:servingstart}). If the actual remaining demand of the target task is no greater than the remaining capacity of the vehicle, then the service is completed successfully, and the target task becomes fully served (lines \ref{line:servingsuccess1}--\ref{line:servingsuccess2}). As soon as the service is completed and the vehicle $k$ becomes idle, the routing policy $h(\cdot)$ is applied to select the next task to be served by the vehicle $k$ (lines \ref{line:servecomplete1}--\ref{line:servecomplete2}). Otherwise, if the service was not successful, a route failure and a collaborative effect occurs. The vehicle partially serves the target task and returns it to the unassigned task set $\Omega_{\text{unass}}$ for any vehicle to potentially complete, before returning to the depot to refill (lines \ref{line:routefailure1}--\ref{line:routefailure2}). This series of actions together constitutes a new recourse policy. 

\begin{algorithm}[!ht]
	\caption{Triggering a serving event.}
	\label{alg:serving}
	\begin{algorithmic}[1]
    	\State get the vehicle $k$ and its current location $v_{\text{curr}}$ in $\epsilon$;
    	\State get the target task $t^*$ to serve;
        \If{$v_{\text{curr}} = v_0$} $q(k) = Q$; \label{line:servingrefill}
        \EndIf
        \If{$v_{\text{curr}} = head(t^*)$} \label{line:servingstart}
            \State $X_{\xi}^{(k)} = (X_{\xi}^{(k)}, tail(t^*))$;
            \If{$\theta(t^*) \cdot d_{\xi}(t^*) \leq q(t)$}
                \State $Y_{\xi}^{(k)} = (Y_{\xi}^{(k)}, \theta(t^*))$; \label{line:servingsuccess1}
                \State $q(k) = q(k) - \theta(t^*) \cdot d_{\xi}(t^*)$, $\theta(t^*) = 0$;
                \State $\Omega_{\text{unser}} = \Omega_{\text{unser}} \setminus t^*$;
                \State $\Gamma = \Gamma \cup$\{serving event for vehicle $k$ to serve $t^*$\}; \label{line:servingsuccess2}
            \Else
                \State $\theta' = q(t) / d_{\xi}(t^*)$; \label{line:routefailure1}
                \State $Y_{\xi}^{(k)} = (Y_{\xi}^{(k)}, \theta')$;
                \State $q(k) = 0$, $\theta(t^*) = \theta(t^*) - \theta'$;
                \State $\Gamma = \Gamma \cup$\{refill event for vehicle $k$\};
                \State $\Omega_{\text{unass}} = \Omega_{\text{unass}} \cup t^*$; \label{line:routefailure2}
            \EndIf
        \ElsIf{$v_{\text{curr}} = tail(t^*)$ and $\theta(t^*) = 0$} \label{line:servecomplete1}
            \State identify the candidate tasks $\bar{\Omega} \subseteq \Omega_{\text{unass}}$; \label{line:cand2}
            \If{$\bar{\Omega} = \emptyset$} \Return;
            \EndIf
            \State select the next task $t^* = \arg\min_{t \in \bar{\Omega}} h(t)$; \label{line:serverule}
            \State $\Gamma = \Gamma \cup$\{serving event for vehicle $k$ to serve $t^*$\}; \label{line:servecomplete2}
        \Else
            \State find the next node $v'$ on the way to $head(t^*)$; \label{line:servego1}
            \State $X_{\xi}^{(k)} = (X_{\xi}^{(k)}, v')$, $Y_{\xi}^{(k)} = (Y_{\xi}^{(k)}, 0)$;
            \State $\Gamma = \Gamma \cup$\{serving event for vehicle $k$ to serve $t^*$\}; \label{line:servego2}
        \EndIf
    \end{algorithmic}
\end{algorithm}

Note that in both Algorithm \ref{alg:refill} (line \ref{line:cand1}) and Algorithm \ref{alg:serving} (line \ref{line:cand2}), a set of candidate tasks $\bar{\Omega}$ is to be identified from the unassigned tasks, out of which the next task is selected by the routing policy. Here, we adopt a simple but effective filter proposed in \citep{Liu2017} where a task's estimated remaining demand, relative to the vehicle's remaining capacity, determines its feasibility. A set of new demand estimation methods for use within this filter are proposed in Sub-section \ref{sec:dem-est}. An unassigned task is considered to be a candidate task if its expected demand is no greater than the remaining capacity of the vehicle, i.e. the task is expected to be served successfully. 

Table \ref{tab:coll} compares the actions of SCPs with and without vehicle collaboration, along with the implementation locations of these collaboration activities in the newly proposed heuristic. Note that the new SCP has the flexibility to accommodate any other type of recourse operator after a route failure.

\begin{table*}[!ht]
    \centering
    \caption{The comparison between Solution Construction Procedures with and without collaboration between vehicles in handling tasks.}
    \label{tab:coll}
    \begin{tabular}{@{}p{2.5cm}p{5cm}p{5cm}@{}}
    \toprule
    {\bf Scenario \& \newline Implementation} & {\bf Without collaboration} & {\bf With collaboration} \\
    \midrule
    Route failure \newline \newline Algorithm \ref{alg:serving}, lines \ref{line:routefailure1}--\ref{line:routefailure2} & The vehicle drives to the depot to refill its capacity, then returns to complete the failed task. & The vehicle returns to the depot to refill its capacity; the partially served task is reintroduced to the unserved task set. \\
    \midrule
    Refill \newline \newline Algorithm \ref{alg:refill}, lines \ref{line:refillcol1}--\ref{line:refillcol2} & To refill, the vehicle returns to the depot to refill directly. & During the refill process, the vehicle tries to (partially) serve the tasks on its way back to the depot, to potentially reduce the workload of other vehicles. \\
    \bottomrule
    \end{tabular}
\end{table*}

\subsection{Estimation of Remaining Demand} 
\label{sec:dem-est}

With collaboration between vehicles, failed tasks are returned to the candidate task set and vehicles can partially serve tasks on their way to refill. As a result, there may be many tasks in $\Omega_{\text{unser}}$ with $\theta (t) < 1$. It is necessary, therefore, to estimate the remaining demand of such partially served tasks. For example, the filter to identify the candidate tasks $\bar{\Omega}$ requires comparing the remaining demand with the remaining capacity. Further, the routing policy may consider the remaining demand as a feature for selecting the next task from the candidate tasks. Note that observing the remaining demand of a partially complete task, be it the actual or an estimated value, is not equivalent to observing the true demand of a task previously unserved. For an unserved task, the expected demand is given by the UCARP instance (i.e. predicted from the historical distribution). However, for a partially served task, it is imprecise to use this expected demand, considering we know that the actual demand must be larger than the amount of demand that has been partially served. 

In the real world, there can be the following two possible scenarios:
\begin{enumerate}
    \item The actual demand of a task becomes exactly known after it is partially served (e.g. a winter road gritting vehicle being able to make an accurate assessment of the amount required to complete a road given the expenditure required so far); 
    \item The actual demand of a task is still unknown after it is partially served and must be estimated (e.g. a waste collection vehicle being unable to exactly define a street's remaining refuse volume given the amount already collected). 
\end{enumerate} 

In the former case, the \emph{actual} remaining demand is known as an accurate estimation. In the latter case, we assume that the random demand follows a normal distribution, which is a commonly adopted assumption. In this case, the remaining demand should follow a \emph{truncated} normal distribution \citep{greene2003econometric}. Specifically, consider a task with a random demand $\tilde{d} \sim \mathcal{N}(\mu_d, \sigma_d)$, after a partial service of the task, a demand of $\Delta_d$ has been served, then the remaining demand after the partial service is:

\begin{equation}
    \mathrm{E}[\tilde{d} | \tilde{d} > \Delta_d] = \mu_d + \sigma_d \frac{\mathtt{pdf}(\alpha)}{1-\mathtt{cdf}(\alpha)}, \label{eq:rd}
\end{equation}

where 
$$
\alpha = \frac{\Delta_d - \mu_d}{\sigma_d},
$$
$$
\mathtt{pdf}(\alpha) = \frac{1}{\sqrt{2\pi}}e^{-\alpha^2 / 2},
$$
$$
\mathtt{cdf}(\alpha) = \frac{1}{2}\left(1+\frac{2}{\sqrt{\pi}}\int_0^{\alpha / \sqrt{2}}e^{-t^2}dt\right).
$$

Obviously, it should be better to use the actual remaining demand whenever possible. However, this is not always available, in which case the approximation based on truncated normal distribution is an alternative. In the experimental studies, we will investigate the effectiveness of such an alternative.

\section{GPHH with Vehicle Collaboration: GPHH-C} \label{sec:GPHH-C}

The proposed heuristic with vehicle collaboration uses a routing policy to make decisions (e.g. line \ref{line:serverule} of Algorithm \ref{alg:serving}). Given the difficulty of manually designing effective routing policies, we propose using GPHH to automate this process. In the proposed GPHH-C (`C'' for collaboration), the SCP with vehicle collaboration is used as the SCP for evaluating the routing policies. Algorithm \ref{alg:gphh} provides the framework of GPHH-C approach. Solutions are generated by passing routing policies to the heuristic which wholly executes the decision making process. During fitness evaluation (line \ref{line:eval}), given a set of training instance samples $\mathbf{I}_{\text{train}}$, the fitness of a routing policy is defined as the average total cost of the solutions obtained by applying this policy to the training samples, i.e.

\begin{equation}
fit(h(\cdot)) = \frac{1}{|\mathbf{I}_{\text{train}}|}\sum_{I_{\xi} \in \mathbf{I}_{\text{train}}} tc(S_{\xi, h(\cdot)}).
\end{equation}

To improve the generalisation of the evolved routing policy, we generate a different set of training samples in each generation (line \ref{line:samples}). Such a training sample rotation strategy has been used in many other studies (e.g. \citep{hildebrandt2010towards,Liu2017}) and shown promise.

\begin{algorithm}[!ht]
	\caption{The framework of GPHH-C}
	\label{alg:gphh}
	\begin{algorithmic}[1]
	    \Require A UCARP instance $I$.
	    \Ensure A routing policy $h^*(\cdot)$.
    	\State Randomly initialise a population of policies;
    	\While{stopping criteria not met}
    	    \State (Re)Generate a set of training samples $\mathbf{I}_{\text{train}}$ of $I$; \label{line:samples}
    	    \ForEach{policy $h(\cdot)$ in the population}
    	        \ForEach{instance sample $I_{\xi} \in \mathbf{I}_{\text{train}}$}
    	            \State generate a solution $S_{\xi, h(\cdot)}$ by Algorithm \ref{alg:colldp}; \label{line:eval}
    	        \EndFor
    	        \State $fit(h(\cdot)) = \frac{1}{|\mathbf{I}_{\text{train}}|}\sum_{I_{\xi} \in \mathbf{I}_{\text{train}}} tc(S_{\xi, h(\cdot)})$; 
    	    \EndFor
    	    Generate the new population;
    	\EndWhile
    	\State \Return the best policy $h^*(\cdot)$ in the final population;
    \end{algorithmic}
\end{algorithm}

\section{Experimental Studies} 
\label{sec:exp}

To verify the effectiveness of the GP-evolved routing policies in the collaborative environment, we compare GPHH-C against two existing models: its non-collaborative counterpart (GPHH); and the EDASLS \citep{Wang2016}, which is the current state-of-the-art algorithm for UCARP.

For each tested algorithm and UCARP instance, the experiment is split into the \emph{training} and \emph{test} phases. During the training phase, a solution (e.g. a routing policy by GPHH \citep{Liu2017} or a robust sequence by EDASLS \citep{Wang2016}) is obtained by using some training samples. Here, we use 5 training samples in each generation.

Then, the trained solution is tested on a set of unseen test instances to show its generalisation. To accurately represent all the possibilities, we generate 500 samples in the test set, and the test performance of a solution is defined as the average total cost over the 500 samples.

\subsection{Dataset}

In the experimental studies, we extend the \emph{gdb}, \emph{val} and \emph{egl} static CARP instances to UCARP instances. The \emph{gdb}, \emph{val} and \emph{egl} datasets are well known CARP instances. The \emph{gdb} instances are mostly small, with at most 55 tasks. The \emph{val} dataset consists of medium sized instances, with the number of tasks ranging from 34 to 97. The \emph{egl} instances are the largest instances, in which the number of tasks can be up to 190. Both the \emph{gdb} and \emph{val} datasets are synthetically generated, whilst the \emph{egl} dataset is is based on a real-world road network from Lancashire, UK. Basic properties of each instance, namely the number of vertices $|V|$, edges $|E|$ and vehicles $m$, can be found in the results Tables \ref{tab:results-ugdb}-\ref{tab:results-uval} below. These are excluded on Table \ref{tab:results-uegl} as for all instances with the \emph{egl-e} prefix $|V| = 77$ and $|E| = 98$ and for all with the \emph{egl-s} prefix $|V| = 140$ and $|E| = 190$. By testing on these instances, we can see the performance of our algorithm on different problem sizes.

For each CARP instance, we transform each task demand $d(t)$ and each traversal cost $\delta_t(e)$ into a random variable $\tilde{d}(t)$ and $\tilde{\delta}_t(e)$. Here, we assume each random variable follows the following normal distribution\footnote{The existing assumption \citep{Mei2010,Wang2016} assumed Gamma distribution with the shape parameter $k = 20$. The resultant Gamma distribution is very close to the normal distribution in this study.}.
$$
\tilde{d}(t) \sim \mathcal{N}\left(d(t), \frac{d(t)}{5}\right), \ \tilde{\delta}_t(e) \sim \mathcal{N}\left(\delta_t(e), \frac{\delta_t(e)}{5}\right).
$$
With no real-world information as guidelines, we set the standard deviation to $20\%$ of the mean (i.e. the static value given in the instance) as a rule of thumb. Note that the random variables can have negative sample values. Here, any negative sampled task demand is modified to $0$, and any negative sampled traversal cost is set to $\infty$ (i.e. the edge becomes inaccessible).

When considering vehicle collaboration, the number of vehicles is an important parameter of Algorithm \ref{alg:colldp}. \citep{Mei2018a} has shown that different routing policies are required for different numbers of vehicles even in the same graph. In this paper, we set the number of vehicles to the minimal required number, i.e. $m = \lceil \sum_{t \in E_T} \text{E}[\tilde{d}(t)] / Q \rceil$, so that each vehicle is expected to have a single trip (no refill).

\subsection{Parameter Settings} \label{sec:exp-params}

All the compared GPHH algorithms share the same parameter settings. Specifically, the terminal set is given in Table \ref{tab:terminals}. The function set is $\{+, -, \times, /, \max, \min\}$. The ``$/$'' operator is protected, and returns $1$ if divided by zero. Table \ref{tab:paraset} shows the parameter settings of the compared algorithms. 

\begin{table}[!ht]
\footnotesize
\begin{center}
\caption{The terminal set used in GPHH and GPHH-C.}
\label{tab:terminals}
\begin{tabular}[!ht] {@{}l@{\hspace{8pt}}p{7.5cm}@{}} 
\toprule
{\bf Notation} & {\bf Description} \\
\midrule
  {CFH} & Cost From Here (the current node) to the candidate task. \\ 
  {CFR1} & Cost From the closest alternative Route to the task. \\
  {CR} & Cost to Refill (from the current node to the depot). \\ 
  {CTD} & Cost from the candidate task To the Depot.  \\ 
  {CTT1} & Cost from the candidate task To its closest unserved Task. \\
  {DEM} & DEMand of the candidate task. \\ 
  {DEM1} & DEMand of the closest unserved task to the candidate task. \\ 
  {FRT} & Fraction of the Remaining (unserved) Tasks . \\ 
  {FUT} & Fraction of the Unassigned Tasks. \\ 
  {FULL} & FULLness of the vehicle (current load over capacity).  \\ 
  {RQ} & Remaining Capacity of the vehicle.  \\ 
  {RQ1} & Remaining Capacity for the closest alternative route.  \\
  {SC} & Serving Cost of the candidate task. \\ 
  {ERC} & a random constant value. \\
\bottomrule
\end{tabular}
\end{center}
\end{table}

\begin{table}[!ht]
\footnotesize
\begin{center}
\caption{The parameter settings for the compared algorithms.}
\label{tab:paraset}
\begin{tabular}[!ht] {@{}cccc@{}} 
\toprule
\multicolumn{2}{c}{\bf GPHH and GPHH-C} &
\multicolumn{2}{c}{\bf EDASLS} \\
\cmidrule(lr){1-2} \cmidrule(lr){3-4}
{\bf Parameter} & {\bf Value} & {\bf Parameter} & {\bf Value} \\
\midrule
Population size & 1024 & Population size & 120 \\
Generations & 51 & Generations & 200 \\
Tournament size & 7 & Tournament size & 7 \\
Crossover rate & 0.8 & Minimal evaluations per generation & 1024 \\
Mutation rate & 0.15 & Local search probability & 0.1 \\
Reproduction rate & 0.05 & \\
Maximal depth & 8 & \\
\bottomrule
\end{tabular}
\end{center}
\end{table}

The parameters of EDASLS were set the same as that in the original literature \citep{Wang2016}. Note that EDASLS does not consider generalisation, i.e. it only optimises the performance on fixed 30 instance samples. To consider generalisation in EDASLS, we divide the EDASLS process into generations, and rotate the training samples after each generation. To be consistent with the GPHH approaches, a generation consists of at least 1024 evaluations (the actual number varies due to the local search in EDASLS).

All the tested algorithms were implemented in Java with the Evolutionary Computation Java library \citep{ecj}. For each UCARP instance, each algorithm was run 30 times independently and their results compared using the Wilcoxon rank sum test with significance level of $0.05$.

\subsection{Results and Discussions}

Tables \ref{tab:results-ugdb}--\ref{tab:results-uegl} show the test performance (average total cost on the 500 test samples) over 30 independent runs. 

In the tables, GPHH-C has two versions, corresponding to the two estimation methods for the remaining demand of the partially served tasks (Section \ref{sec:dem-est}). The column ``Actual'' stands for vehicles knowing the actual remaining demand, whereas ``Truncate'' means using the truncated normal distribution to estimate the remaining demand. The results of EDASLS and GPHH are associated with two markers ``+/-/='', indicating for the statistical comparison results with the two GPHH-C versions using the Wilcoxon rank sum test with significance level of $0.05$. The first one is with ``Actual'', and the second one is with ``Truncate''. ``+'', ``-'' and ``='' indicate that the results of the algorithm are statistically significantly higher (worse) than, lower (better) than, and comparable to the results of the corresponding GPHH-C results. For example, on \emph{Ugdb}11 in Table \ref{tab:results-ugdb}, EDASLS outperforms both GPHH-C algorithms, whilst GPHH is significantly worse than GPHH-C \emph{Actual} and is comparable to GPHH-C \emph{Truncate}. In addition, we compare between the two demand estimation methods of GPHH-C. If either of them is significantly lower (better), then the corresponding result is marked in bold.

\begin{table*} 
	\footnotesize
	\centering
	\caption{The test performance of the EDASLS, GPHH, and GPHH-C (with actual and truncate demand estimation) on the \emph{ugdb} instances.}
	\label{tab:results-ugdb}
	\begin{tabular} { @{} c  c  c  c  c  c  c  c  c @{} } 
		\toprule
		& & & & & & \multicolumn{2}{c}{GPHH-C} \\ 
		\cmidrule(lr){7-8}
		Instance & $|V|$ & $|E|$ & $m$ & EDASLS & GPHH & Actual & Truncate \\
		\midrule
		gdb1 & 12 & 22 & 5 & 338.25(0.14){\bf (+)}{\bf (+)} & 337.54(2.64){\bf (+)}{\bf (+)}  & 330.25(2.60) & 330.44(2.95) \\
		gdb2 & 12 & 26 & 6 & 367.86(0.44){\bf (+)}{\bf (+)} & 369.00(5.40){\bf (+)}{\bf (+)}  & 363.81(4.89) & {\bf 360.70(3.83)} \\
		gdb3 & 12 & 22 & 5 & 298.21(0.31){\bf (+)}{\bf (+)} & 297.18(3.58){\bf (=)}{\bf (=)}  & 297.66(7.55) & 295.83(6.64) \\
		gdb4 & 11 & 19 & 4 & 314.02(1.15){\bf (+)}{\bf (+)} & 323.95(4.53){\bf (+)}{\bf (+)}  & 309.99(5.28) & 310.83(6.21) \\
		gdb5 & 13 & 26 & 6 & 410.36(0.27){\bf (+)}{\bf (+)} & 415.47(2.97){\bf (+)}{\bf (+)}  & 404.77(5.16) & 405.00(5.86) \\
		gdb6 & 12 & 22 & 5 & 324.99(0.17){\bf (-)}{\bf (-)} & 339.61(8.79){\bf (=)}{\bf (=)}  & 340.20(5.14) & 339.63(7.96) \\
		gdb7 & 12 & 22 & 5 & 352.87(1.45){\bf (+)}{\bf (+)} & 351.32(13.28){\bf (+)}{\bf (+)}  & 339.16(7.10) & 338.18(3.82) \\
		gdb8 & 27 & 46 & 10 & 449.16(10.10){\bf (+)}{\bf (+)} & 448.09(12.49){\bf (+)}{\bf (+)}  & 429.61(12.53) & 428.68(13.01) \\
		gdb9 & 27 & 51 & 10 & 373.47(6.86){\bf (+)}{\bf (+)} & 377.09(9.32){\bf (+)}{\bf (+)}  & 367.45(5.58) & 368.52(8.52) \\
		gdb10 & 12 & 25 & 4 & 283.64(0.77){\bf (-)}{\bf (-)} & 296.37(6.20){\bf (+)}{\bf (+)}  & 291.36(10.03) & 290.93(6.04) \\
		gdb11 & 22 & 45 & 5 & 415.91(6.25){\bf (-)}{\bf (-)} & 423.22(2.50){\bf (+)}{\bf (=)}  & 422.50(5.74) & 424.65(8.07) \\
		gdb12 & 13 & 23 & 7 & 533.09(11.29){\bf (-)}{\bf (-)} & 614.27(22.06){\bf (+)}{\bf (+)}  & {\bf 591.88(19.55)} & 600.17(17.81) \\
		gdb13 & 10 & 28 & 6 & 565.96(3.71){\bf (-)}{\bf (-)} & 571.45(3.29){\bf (+)}{\bf (+)}  & 569.66(3.68) & 569.64(5.24) \\ 
		gdb14 & 7 & 21 & 5 & 104.15(0.08){\bf (=)}{\bf (=)} & 105.69(1.69){\bf (=)}{\bf (=)}  & 105.24(1.84) & 105.41(1.92) \\
		gdb15 & 7 & 21 & 4 & 60.08(0.08){\bf (+)}{\bf (+)} & 58.13(0.12){\bf (=)}{\bf (-)}  & 58.43(1.82) & 58.15(0.42) \\
		gdb16 & 8 & 28 & 5 & 131.25(0.29){\bf (-)}{\bf (-)} & 133.60(1.59){\bf (=)}{\bf (=)}  & 133.14(1.49) & 133.34(1.68) \\
		gdb17 & 8 & 28 & 5 & 91.14(0.15){\bf (-)}{\bf (-)} & 91.16(0.09){\bf (=)}{\bf (=)}  & 91.21(0.16) & 91.20(0.17) \\
		gdb18 & 9 & 36 & 5 & 171.27(1.10){\bf (+)}{\bf (+)} & 168.49(2.15){\bf (+)}{\bf (+)}  & 167.78(3.23) & 167.39(2.95) \\
		gdb19 & 8 & 11 & 3 & 63.12(0.06){\bf (+)}{\bf (+)} & 61.93(2.22){\bf (+)}{\bf (+)}  & 60.77(1.30) & 61.01(1.40) \\
		gdb20 & 11 & 22 & 4 & 125.18(0.33){\bf (-)}{\bf (-)} & 128.48(1.08){\bf (+)}{\bf (=)}  & 127.67(1.40) & 128.06(1.82) \\
		gdb21 & 11 & 33 & 6 & 161.22(0.91){\bf (-)}{\bf (-)} & 163.45(1.60){\bf (+)}{\bf (+)}  & 162.85(3.33) & 162.18(1.45) \\
		gdb22 & 11 & 44 & 8 & 208.25(0.91){\bf (-)}{\bf (-)} & 210.29(1.81){\bf (=)}{\bf (=)}  & 209.94(2.29) & 209.91(2.62) \\
		gdb23 & 11 & 55 & 10 & 248.82(1.62){\bf (-)}{\bf (-)} & 251.04(2.43){\bf (=)}{\bf (=)}  & 250.91(2.47) & 250.34(2.10) \\ 
		\midrule
		average & & & & 277.92 & 284.21 & 279.40 & 279.57 \\
		\bottomrule
	\end{tabular}
\end{table*}

\begin{table*}[!ht]
	\footnotesize
	\centering
	\caption{The test performance of the EDASLS, GPHH, and GPHH-C (with actual and truncate demand estimation) on the \emph{uval} instances.}
	\label{tab:results-uval}
	\begin{tabular} { @{} c  c  c  c  c  c  c  c  c @{} } 
		\toprule
		& & & & & & \multicolumn{2}{c}{GPHH-C} \\ 
		\cmidrule(lr){7-8}
		Instance & $|V|$ & $|E|$ & $m$ & EDASLS & GPHH & Actual & Truncate \\
		\midrule
		val1A & 24 & 39 & 2 & 172.97(0.01){\bf (-)}{\bf (-)} & 176.52(2.58){\bf (=)}{\bf (=)}  & 175.90(3.22) & 175.72(3.08) \\
		val1B & 24 & 39 & 3 & 187.31(1.68){\bf (+)}{\bf (+)} & 184.95(2.65){\bf (=)}{\bf (=)}  & 184.24(1.43) & 184.80(2.25) \\
		val1C & 24 & 39 & 8 & 293.24(5.52){\bf (-)}{\bf (-)} & 313.55(8.87){\bf (+)}{\bf (+)}  & 302.60(9.21) & 303.27(7.25) \\
		val2A & 24 & 34 & 2 & 241.34(8.64){\bf (+)}{\bf (+)} & 230.14(2.64){\bf (=)}{\bf (=)}  & 229.64(3.07) & 229.44(2.23) \\
		val2B & 24 & 34 & 3 & 281.19(5.43){\bf (+)}{\bf (+)} & 277.81(3.70){\bf (+)}{\bf (=)}  & 275.55(2.78) & 276.69(3.67) \\
		val2C & 24 & 34 & 8 & 586.68(12.13){\bf (+)}{\bf (+)} & 593.59(15.37){\bf (+)}{\bf (+)}  & 556.98(15.78) & 554.93(15.06) \\
		val3A & 24 & 35 & 2 & 84.56(1.48){\bf (+)}{\bf (+)} & 81.85(0.77){\bf (=)}{\bf (=)}  & 82.18(1.54) & 81.91(0.90) \\
		val3B & 24 & 35 & 3 & 95.29(1.42){\bf (=)}{\bf (+)} & 95.86(3.49){\bf (=)}{\bf (=)}  & 95.15(1.78) & 94.39(1.74) \\
		val3C & 24 & 35 & 7 & 176.57(5.06){\bf (+)}{\bf (+)} & 175.90(7.25){\bf (+)}{\bf (+)}  & 165.49(5.48) & 167.44(7.46) \\
		val4A & 41 & 69 & 3 & 413.51(2.73){\bf (-)}{\bf (-)} & 419.56(6.80){\bf (=)}{\bf (=)}  & 421.14(12.10) & 419.17(5.83) \\
		val4B & 41 & 69 & 4 & 434.46(5.88){\bf (-)}{\bf (-)} & 448.27(10.91){\bf (=)}{\bf (=)}  & 444.33(8.35) & 443.81(7.06) \\
		val4C & 41 & 69 & 5 & 480.13(4.99){\bf (=)}{\bf (-)} & 492.20(11.85){\bf (+)}{\bf (+)}  & 484.81(11.04) & 487.07(7.96) \\
		val4D & 41 & 69 & 9 & 649.36(9.85){\bf (-)}{\bf (-)} & 701.20(32.64){\bf (+)}{\bf (+)}  & 670.09(22.12) & 678.73(21.59)\\
		val5A & 34 & 65 & 3 & 447.53(3.18){\bf (+)}{\bf (+)} & 441.30(4.55){\bf (=)}{\bf (=)}  & 441.66(6.38) & 441.07(3.62) \\
		val5B & 34 & 65 & 4 & 478.37(3.60){\bf (+)}{\bf (+)} & 469.99(5.98){\bf (=)}{\bf (=)}  & 467.45(4.90) & 469.54(5.25) \\
		val5C & 34 & 65 & 5 & 535.41(8.93){\bf (+)}{\bf (+)} & 517.87(9.97){\bf (+)}{\bf (+)}  & 511.19(16.78) & 507.23(8.76) \\
		val5D & 34 & 65 & 9 & 723.28(8.46){\bf (+)}{\bf (+)} & 726.79(13.53){\bf (+)}{\bf (+)}  & 697.13(8.96) & 697.68(11.08) \\
		val6A & 31 & 50 & 3 & 229.48(1.24){\bf (+)}{\bf (+)} & 228.31(2.08){\bf (=)}{\bf (=)}  & 227.87(1.49) & 227.93(2.01) \\
		val6B & 31 & 50 & 4 & 254.88(3.91){\bf (=)}{\bf (=)} & 257.98(3.46){\bf (+)}{\bf (+)}  & 256.03(3.83) & 256.06(2.91) \\
		val6C & 31 & 50 & 10 & 383.49(6.82){\bf (-)}{\bf (-)} & 398.94(10.17){\bf (+)}{\bf (+)}  & 392.85(12.65) & 392.35(10.90) \\
		val7A & 40 & 66 & 3 & 279.95(1.46){\bf (-)}{\bf (-)} & 287.15(3.39){\bf (=)}{\bf (=)}  & 287.43(4.07) & 287.12(4.32) \\
		val7B & 40 & 66 & 4 & 284.60(2.02){\bf (-)}{\bf (-)} & 299.07(10.13){\bf (=)}{\bf (+)}  & 294.79(8.23) & 292.89(7.99) \\
		val7C & 40 & 66 & 9 & 386.58(5.08){\bf (-)}{\bf (-)} & 421.18(15.89){\bf (=)}{\bf (=)}  & 414.06(11.32) & 421.42(18.61) \\
		val8A & 30 & 63 & 3 & 396.02(2.38){\bf (-)}{\bf (-)} & 400.60(4.04){\bf (=)}{\bf (+)}  & 398.83(3.14) & 397.51(3.54) \\
		val8B & 30 & 63 & 4 & 430.66(5.97){\bf (+)}{\bf (+)} & 426.93(8.19){\bf (+)}{\bf (+)}  & 422.88(8.57) & 420.02(4.70) \\
		val8C & 30 & 63 & 9 & 672.24(13.25){\bf (+)}{\bf (+)} & 667.96(14.09){\bf (+)}{\bf (+)}  & 650.25(12.23) & 649.05(13.07) \\
		val9A & 50 & 92 & 3 & 331.33(2.66){\bf (-)}{\bf (-)} & 335.97(1.83){\bf (=)}{\bf (=)}  & 335.64(3.03) & 336.09(5.11) \\
		val9B & 50 & 92 & 4 & 345.53(4.54){\bf (=)}{\bf (=)} & 348.18(2.98){\bf (=)}{\bf (=)}  & 347.35(2.49) & 346.81(3.38) \\
		val9C & 50 & 92 & 5 & 365.65(4.31){\bf (+)}{\bf (+)} & 362.58(3.50){\bf (=)}{\bf (=)}  & 361.26(4.98) & 361.61(4.35) \\
		val9D & 50 & 92 & 10 & 477.10(6.21){\bf (+)}{\bf (+)} & 470.84(5.97){\bf (+)}{\bf (+)}  & {\bf 458.05(5.57)} & 461.87(4.76) \\
		val10A & 50 & 97 & 3 & 442.85(2.67){\bf (=)}{\bf (=)} & 443.66(4.59){\bf (=)}{\bf (=)}  & 442.77(4.40) & 441.73(2.61) \\
		val10B & 50 & 97 & 4 & 456.81(4.39){\bf (=)}{\bf (=)} & 458.03(7.04){\bf (=)}{\bf (=)}  & 455.90(5.38) & 455.05(4.39) \\
		val10C & 50 & 97 & 5 & 483.49(6.61){\bf (+)}{\bf (+)} & 479.55(6.39){\bf (+)}{\bf (=)}  & 477.04(7.53) & 476.61(8.31) \\
		val10D & 50 & 97 & 10 & 627.17(7.03){\bf (+)}{\bf (+)} & 616.48(9.98){\bf (+)}{\bf (+)}  & 595.63(5.74) & 596.81(6.06) \\ 
		\midrule
		average & & & & 386.15 & 389.73 & 383.06 & 383.35 \\
		\bottomrule
	\end{tabular}
\end{table*}

\begin{table*}[!ht]
	\footnotesize
	\centering
	\caption{The test performance of the EDASLS, GPHH, and GPHH-C (with actual and truncate demand estimation) on the \emph{uegl} instances. All instances with the \emph{egl-e} prefix: $|V| = 77$ and $|E| = 98$; \emph{egl-s} prefix: $|V| = 140$ and $|E| = 190$.}
	\label{tab:results-uegl}
	\begin{tabular} { @{} c  c  c  c  c  c  c  c @{} } 
		\toprule
		& & & & & \multicolumn{2}{c}{GPHH-C} \\ 
		\cmidrule(lr){6-7}
		Instance & $|E_T|$ & $m$ & EDASLS & GPHH & Actual & Truncate \\
		\midrule
		egl-e1-A & 51 & 5 & 4302.24(64.79){\bf (+)}{\bf (+)} & 4454.61(99.53){\bf (+)}{\bf (+)}  & 4241.02(86.98) & 4239.13(88.29) \\
		egl-e1-B & 51 & 7 & 5559.52(99.97){\bf (+)}{\bf (+)} & 5584.20(155.43){\bf (+)}{\bf (+)}  & 5316.84(96.13) & 5323.03(114.24) \\
		egl-e1-C & 51 & 10 & 7272.74(111.79){\bf (+)}{\bf (+)} & 7245.36(166.21){\bf (+)}{\bf (+)}  & 6913.57(147.94) & 6981.05(152.48) \\
		egl-e2-A & 72 & 7 & 6190.26(112.32){\bf (+)}{\bf (+)} & 6396.96(166.56){\bf (+)}{\bf (+)}  & 6012.11(80.59) & 6056.04(88.12) \\
		egl-e2-B & 72 & 10 & 8250.08(191.17){\bf (+)}{\bf (+)} & 8264.63(269.36){\bf (+)}{\bf (+)}  & 7853.43(164.90) & 7880.55(196.75) \\
		egl-e2-C & 72 & 14 & 11433.16(210.83){\bf (+)}{\bf (+)} & 10804.76(303.43){\bf (+)}{\bf (+)}  & 10282.27(291.27) & 10368.49(384.72) \\
		egl-e3-A & 87 & 8 & 7495.17(177.51){\bf (+)}{\bf (+)} & 7697.28(401.47){\bf (+)}{\bf (+)}  & 7193.68(141.65) & 7212.97(122.04) \\
		egl-e3-B & 87 & 12 & 10743.42(219.90){\bf (+)}{\bf (+)} & 10306.60(193.44){\bf (+)}{\bf (+)}  & 9824.36(197.43) & 9812.44(209.56) \\
		egl-e3-C & 87 & 17 & 14203.53(274.56){\bf (+)}{\bf (+)} & 13518.83(218.27){\bf (+)}{\bf (+)}  & 12706.95(250.83) & 12717.20(253.25) \\
		egl-e4-A & 98 & 9 & 8375.67(176.82){\bf (+)}{\bf (+)} & 8256.50(498.75){\bf (+)}{\bf (+)}  & 7994.38(387.96) & 7924.83(469.76) \\
		egl-e4-B & 98 & 14 & 12177.31(205.99){\bf (+)}{\bf (+)} & 11808.94(267.26){\bf (+)}{\bf (+)}  & 11208.18(250.03) & 11192.95(188.72) \\
		egl-e4-C & 98 & 19 & 15502.01(249.29){\bf (+)}{\bf (+)} & 15327.12(361.94){\bf (+)}{\bf (+)}  & 14352.95(218.77) & 14303.85(309.97) \\
		egl-s1-A & 75 & 7 & 6624.66(110.15){\bf (+)}{\bf (+)} & 6706.60(205.98){\bf (+)}{\bf (+)}  & 6549.26(207.34) & 6569.38(230.12) \\
		egl-s1-B & 75 & 10 & 8895.93(153.39){\bf (+)}{\bf (+)} & 8668.43(195.03){\bf (+)}{\bf (+)}  & 8443.71(215.35) & 8403.55(152.23) \\
		egl-s1-C & 75 & 14 & 11665.94(271.12){\bf (+)}{\bf (+)} & 12013.78(263.44){\bf (+)}{\bf (+)}  & 11398.01(195.84) & 11423.03(231.77) \\
		egl-s2-A & 147 & 14 & 13817.53(202.27){\bf (+)}{\bf (+)} & 13129.83(556.42){\bf (+)}{\bf (+)}  & 12568.43(469.06) & 12623.58(498.73) \\
		egl-s2-B & 147 & 20 & 19226.22(322.84){\bf (+)}{\bf (+)} & 17902.05(417.37){\bf (+)}{\bf (+)}  & 17112.74(405.98) & 17151.63(343.84) \\
		egl-s2-C & 147 & 27 & 24504.86(455.05){\bf (+)}{\bf (+)} & 23299.85(370.32){\bf (+)}{\bf (+)}  & 21774.97(401.79) & 21872.97(357.30) \\
		egl-s3-A & 159 & 15 & 14280.23(242.63){\bf (+)}{\bf (+)} & 13830.32(540.00){\bf (+)}{\bf (+)}  & 13507.25(367.78) & 13549.47(387.07) \\
		egl-s3-B & 159 & 22 & 20228.78(308.71){\bf (+)}{\bf (+)} & 19192.70(310.52){\bf (+)}{\bf (+)}  & 18290.47(368.07) & 18394.26(587.83) \\
		egl-s3-C & 159 & 29 & 26032.72(584.60){\bf (+)}{\bf (+)} & 24742.34(448.14){\bf (+)}{\bf (+)}  & 23335.50(439.46) & 23452.58(482.04) \\
		egl-s4-A & 190 & 19 & 17572.98(237.64){\bf (+)}{\bf (+)} & 17229.92(281.10){\bf (+)}{\bf (+)}  & 16586.74(304.57) & 16717.33(379.17) \\
		egl-s4-B & 190 & 27 & 24504.42(344.10){\bf (+)}{\bf (+)} & 23334.82(497.00){\bf (+)}{\bf (+)}  & 22049.57(460.45) & 22185.40(514.91) \\
		egl-s4-C & 190 & 35 & 31677.90(865.15){\bf (+)}{\bf (+)} & 30139.56(568.08){\bf (+)}{\bf (+)}  & 28034.04(452.61) & 28213.56(476.70) \\ 
		\midrule
		average & & & 13772.39 & 13327.33 & 12647.94 & 12690.39 \\
		\bottomrule
	\end{tabular}
\end{table*}

From the tables, we have the following observations.
\begin{enumerate}
    \item There was no statistical significance between the two versions of GPHH-C on most test instances. The only three exceptions out of the total 81 instances are \emph{Ugdb}2, \emph{Ugdb}12 and \emph{Uval}9D. This is a very promising pattern, which indicates that we do not require the assumption of knowing the actual remaining demand, and the truncated normal distribution estimation can achieve almost the same performance.
    \item On the \emph{Ugdb} dataset, GPHH-C had a mixed relationship with EDASLS. Both versions performed significantly better than EDASLS on 11 instances, and significantly worse on 11 instances (tie on \emph{Ugdb}14). Both GPHH-C versions significantly outperformed GPHH on 13 \emph{Ugdb} instances, and were defeated by GPHH only on 1 instance (\emph{Ugdb}15). On average, EDASLS performed better than GPHH and GPHH-C on the small and simple \emph{Ugdb} dataset. This is likely due to the low cost of route failure recourse relative to the cost of the solution: on small instances, returning to the depot to refill is rarely prohibitive.
    \item On the \emph{Uval} dataset, GPHH-C obtained better performance in comparison to EDASLS. Both versions significantly outperformed EDASLS on at least 17 instances, while GPHH-C \emph{Truncate} performed significantly worse than EDASLS on 12 instances. The relative relationship between GPHH and GPHH-C showed the same pattern. Both GPHH-C versions performed significantly better than GPHH on 13 instances, but never showed significantly worse performance. Note that GPHH-C significantly outperformed GPHH on most instances with a large number of vehicles (e.g. the C and D suffix instances). This is consistent with our expectation, as there should be more opportunity for vehicle collaboration as the number of vehicles increases. The average performance of EDASLS was slightly better than GPHH (386.15 versus 389.73), but was beaten by GPHH-C (383.06 and 383.35). 
    \item GPHH-C showed the most obvious advantage on the \emph{Uegl} dataset, with both versions significantly outperforming EDASLS and GPHH on all instances. The performance of GPHH became better than EDASLS on this dataset (13327.33 versus 13772.39), as well. This demonstrates that the reactive approaches show a more obvious advantage over the proactive approaches (which maintain a robust solution) for large and complex instances. This is likely due to the fact that the \emph{solution} search space explored in EDASLS scales with instance size, whilst the \emph{heuristic} search space explored in GPHH remains static. Further, the cost benefit of collaborative policies is highlighted most when vehicles are able to avoid large recourse cost trips back to the partially complete task, as indicated by the high performance of GPHH-C.   
\end{enumerate}

\begin{figure}[t]
    \begin{minipage}{0.48\textwidth}
        \centering
        \footnotesize
        \includegraphics[width=\textwidth]{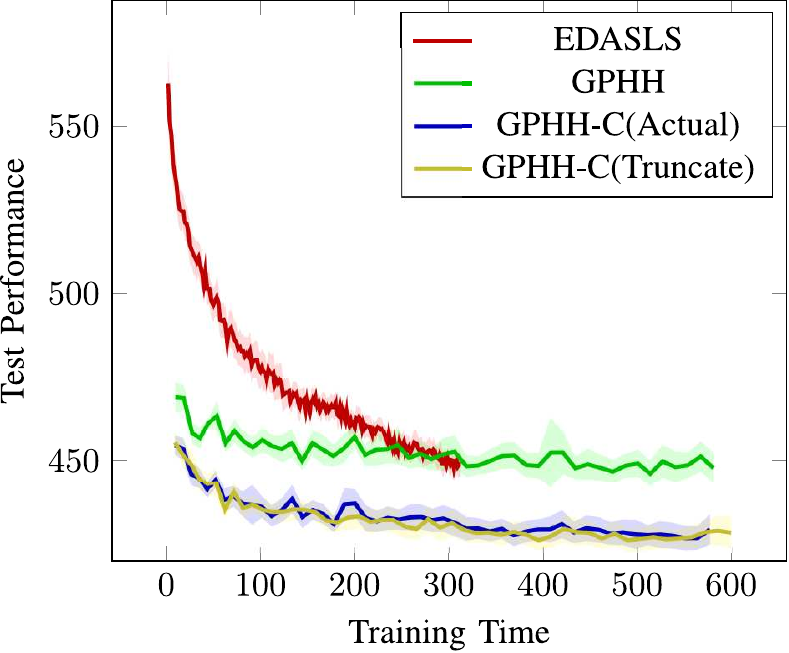}
        \caption{The convergence curves of compared algorithms on \emph{Ugdb}8.}
        \label{fig:curve-gdb8}
    \end{minipage}\hfill
    \begin{minipage}{0.48\textwidth}
        \centering
        \footnotesize
        \includegraphics[width=\textwidth]{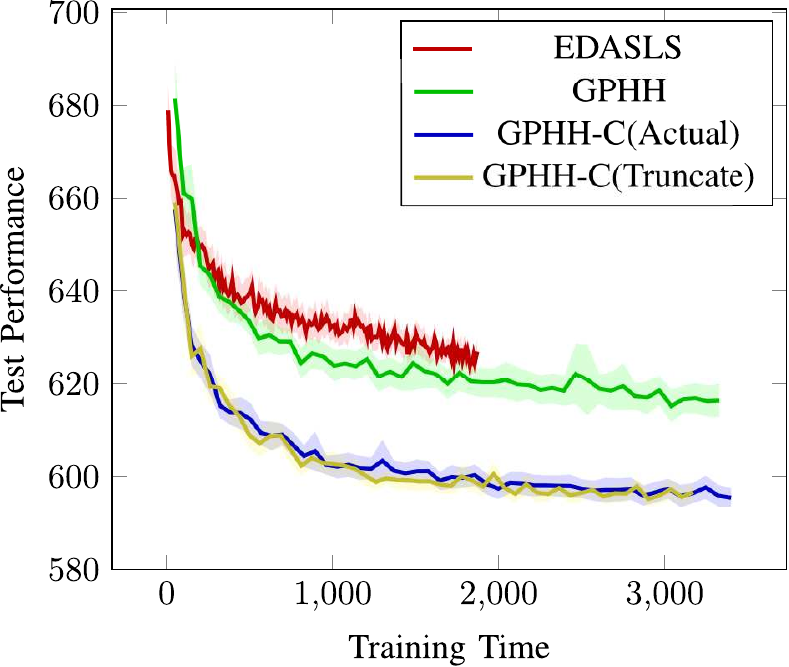}
        \caption{The convergence curves of compared algorithms on \emph{Uval}10D.}
        \label{fig:curve-val10D}
    \end{minipage}\hfill \vfill
    \centering
    \begin{minipage}{0.48\textwidth}
        \centering
        \footnotesize
        \includegraphics[width=\textwidth]{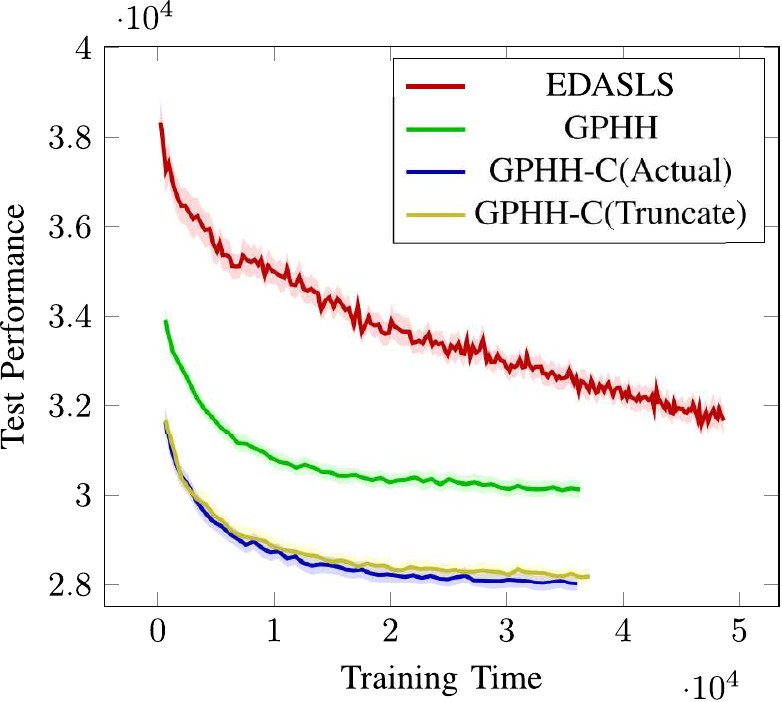}
        \caption{The convergence curves of compared algorithms on \emph{Uegl}-s4-C.}
        \label{fig:curve-egl-s4-c}
        \end{minipage}\hfill 
\end{figure}

Figs. \ref{fig:curve-gdb8}--\ref{fig:curve-egl-s4-c} show the convergence curves of the compared algorithms on three representative instances (one from each dataset), where the $x$-axis stands for the training time, and the $y$-axis is the  test performance (averaged over 30 independent runs) of the best-so-far solution or routing policy during the training process. The ribbon indicates the 95\% confidence interval. Both GPHH-C methods significantly outperformed the EDASLS and GPHH approaches in these three examples.   

From \emph{Ugdb8} in Fig. \ref{fig:curve-gdb8}, we can see that the curve of EDASLS starts significantly higher than that of the GPHH approaches and rapidly converges to compete with GPHH. Both GPHH-C curves, however, are always significantly below both of these techniques. This is expected, as the benefits of collaboration are available from the outset via the decision making process, independent of the quality of the policy. High performing policies simply better exploit the collaborative environment. The training time of EDASLS was much shorter than the GPHH approaches as the fitness evaluation of EDASLS is much less time consuming than the decision making process used in GPHH. 

In \emph{Uval}10D, shown in Fig. \ref{fig:curve-val10D}, we further see the advantage EDASLS has with regards to computation time. However, despite early competition, the performance of EDASLS fails to match that of the GPHH approach in convergence. Continuing the pattern shown in Fig. \ref{fig:curve-gdb8}, the GPHH-C approaches constantly outperform the alternative approaches by a significant margin. 

In the largest and most complex \emph{Uegl}-s4-C instance, the compared algorithms showed very different performance. Fig. \ref{fig:curve-egl-s4-c} clearly shows the advantage GPHH methods have over EDASLS, and specifically the much stronger performance of GPHH-C over its precursor. Note that in \emph{Uegl}-s4-C, the training time of EDASLS (with the same 200 generations) is much longer than that of the GPHH-based approaches. The reason is due to the local search in EDASLS. The complexity of local search used in EDASLS is $O(n^2)$, where $n = |E_T|$. Therefore, it is highly likely that even one step of local search induces much more fitness evaluations than the threshold (i.e. 1024), especially for large instances. For example, on average over the 30 runs, EDASLS spent 1468 evaluations per generation solving \emph{Ugdb}8 (46 tasks), 2185 solving \emph{Uval}10D (97 tasks) and 11,790 solving \emph{Uegl}-s4-C (190 tasks). This sharp increase is attributable to the complexity of the local search in EDASLS having an exponential relationship with problem size (i.e. with the number of required tasks). This further highlights the scalability benefits of the GPHH approach over EDASLS.


Fig. \ref{fig:curve-gdb8} shows EDASLS may not have converged on \emph{Ugdb}8. Running the algorithm for the same time as the GPHH methods improves performance considerably, such to compete with GPHH-C. Conversely, extending EDASLS out to 1000 generations on \emph{Uegl}-s4-C, whilst taking \emph{significantly} longer to compute, improves performance, however only such to compete with GPHH, not GPHH-C.

\begin{figure}
    \centering
    \begin{minipage}{0.6\textwidth}
        \centering
        \begin{tikzpicture}
            \begin{axis}[
                width=\textwidth,
                enlargelimits=false,
                xmax=200,
                ymax=35,
                ylabel=Number of vehicles,
         	    xlabel=Number of required tasks
                ]
        
              \addplot[thick,blue] graphics[xmin=0,ymin=0,xmax=200,ymax=35] {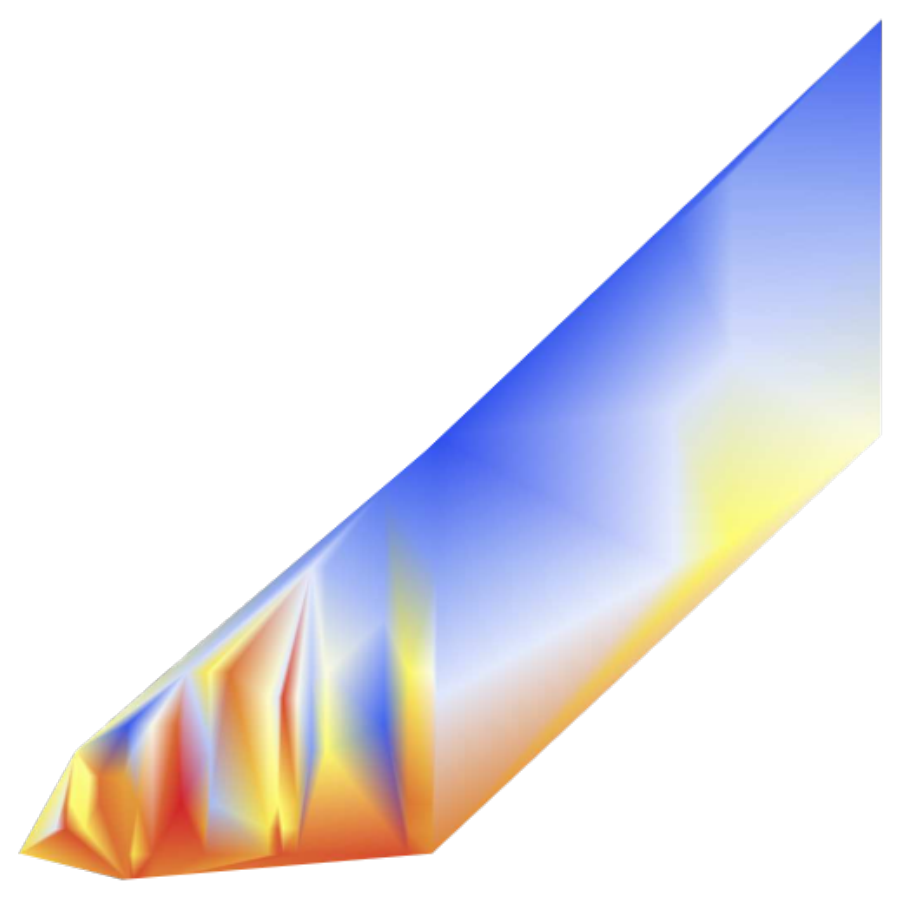};
            \end{axis}
        \end{tikzpicture}
    \end{minipage}
    \begin{minipage}{0.15\textwidth}
        \centering
        \vspace*{-1.1cm}\hspace*{-1cm}\includegraphics[width=0.55\textwidth]{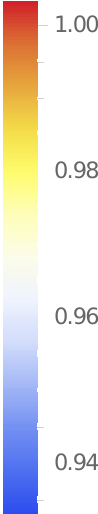} 
    \end{minipage}
    
    \caption{The ratio between the test performances of GPHH-C and GPHH versus the numbers of tasks and vehicles.}
    \label{fig:heatMap}
\end{figure}

Fig. \ref{fig:heatMap} shows the heat map of the ratio between the test performances of the GPHH-C and GPHH versus the numbers of tasks and vehicles. The heat map was drawn based on the 81 points obtained on the 81 instances. A smaller ratio indicates that GPHH-C shows a more obvious advantage over GPHH (ratio $< 1$ means GPHH-C is better than GPHH). The figure clearly shows that the ratio is almost always smaller than 1. More importantly, there is an obvious trend that the ratio decreases as the number of tasks and vehicles increases.

In summary, we can see obvious advantage of GPHH-C over both EDASLS and GPHH, especially on the complex UCARP instances with a large number of tasks and vehicles. This demonstrates the continued feasibility of GPHH as a solution technique for UCARP and the effectiveness of vehicle collaboration in solving UCARP using routing policies.  

\section{Further Analysis}
\label{sec:ana}

\subsection{Effectiveness of the Heuristic with Collaboration}

It has been shown that GPHH-C can obtain solutions with much lower total cost in real time. The quality of a solution depends on (1) the SCP that generates the solution and (2) the routing policy. In order to investigate the effectiveness of the newly proposed heuristic with vehicle collaboration independently, we tested five manually designed path scanning routing policies \citep{Lacomme2004} in the heuristics with and without collaboration. Specifically, the priority functions of the five routing policies are described in Table \ref{tab:ps}, where $\alpha$ is set to a sufficiently large value (10000 in this case) to guarantee the priority of the CFH terminal (Table \ref{tab:terminals}), i.e. they always select among the nearest neighbours.

\begin{table}[!ht]
    \footnotesize
    \centering
    \caption{The manually designed routing policies \citep{Lacomme2004}}
    \label{tab:ps}
    \begin{tabular}{@{}cc@{}}
        \toprule
        Manually Designed Policy  & $h(\cdot)$ \\
        \midrule
        PS1  & $\alpha \times \text{CFH} - \text{CTD}$ \\
        PS2  & $\alpha \times \text{CFH} + \text{CTD}$ \\
        PS3  & $\alpha \times \text{CFH} - \text{DEM} / \text{SC}$ \\
        PS4  & $\alpha \times \text{CFH} + \text{DEM} / \text{SC}$ \\
        PS5  & uses PS1 if $\text{FULL} < 0.5$, and PS2 otherwise. \\
        \bottomrule
    \end{tabular}
\end{table}

Table \ref{tab:psresults} shows the average test performance of PS1--PS5 with (w/) and without (w/o) collaboration on the \emph{Ugdb}, \emph{Uval} and \emph{Uegl} datasets. From the table, it is clear that for all the five routing policies, the heuristic with collaboration generated much better solutions. A deeper look shows that among all the $5 \times 81 = 405$ comparisons, the heuristics with collaboration showed better test performance than their counterpart without collaboration in all but 7 comparisons. 

\begin{table*}[!ht]
    \footnotesize
    \centering
    \caption{The average test performance of PS1--PS5 with (w/) and without (w/o) collaboration (c) on the \emph{Ugdb}, \emph{Uval} and \emph{Uegl} datasets.}
    \label{tab:psresults}
    \begin{tabular}{@{}ccccccccccc@{}}
        \toprule
         & \multicolumn{2}{c}{PS1} & \multicolumn{2}{c}{PS2} & \multicolumn{2}{c}{PS3} & \multicolumn{2}{c}{PS4} & \multicolumn{2}{c}{PS5} \\
        \cmidrule(lr){2-3}
        \cmidrule(lr){4-5}
        \cmidrule(lr){6-7}
        \cmidrule(lr){8-9}
        \cmidrule(lr){10-11}
        & w/o c & w/ c & w/o c & w/ c & w/o c & w/ c & w/o c & w/ c & w/o c & w/ c \\
        \midrule
        \emph{Ugdb} & 324.1 & \textbf{321.2} & 356.6	& \textbf{350.8} & 335.9 & \textbf{332.7} & 342.4 & \textbf{337.3} & 323.4 & \textbf{320.3} \\
        \emph{Uval} & 441.6 & \textbf{434.0} & 	507.2 & \textbf{494.6} & 474.5 & \textbf{466.5} & 473.5 & \textbf{463.0} & 	476.5	& \textbf{468.3} \\
        \emph{Uegl} & 17506.6 & \textbf{16489.9} & 17465.8 & \textbf{16470.9} & 17473.2 & \textbf{16486.6} & 17480.3 & \textbf{16459.9} & 17526.6 & \textbf{16554.2} \\
        \bottomrule
    \end{tabular}
\end{table*}

\subsection{Effectiveness of Collaboration Components}

As mentioned in Section \ref{sec:dp} and Table \ref{tab:coll}, the proposed GPHH-C consists of two types of collaborations, one during route failure, and the other during refill. To investigate the effectiveness of each type of collaboration, we compare GPHH-C with its variants with only the route failure collaboration (namely GPHH-C$_\text{RouteFailure}$) and only the refill collaboration (namely GPHH-C$_\text{Refill}$), respectively. For the sake of comprehensive comparison, EDASLS and GPHH were also included in the comparison.

Table \ref{tab:collres} shows the average test performance of the GPHH-C with different collaboration components (with truncated demand estimation), and of EDASLS and GPHH. Table \ref{tab:wdlEx} gives the win-draw-lose results of the pairwise comparisons between the algorithms, i.e. the number of instances among the total 81 instances that an algorithm performed significantly better than (win), statistically the same as (draw), and worse than (lose) the other algorithm. 

First, comparing the results of the baseline GPHH in Table \ref{tab:collres} against those of the manually designed heuristics in Table \ref{tab:psresults} highlights the benefit of utilising automatic routing policy design. Second, from the table, one can see that both GPHH-C$_\text{RouteFailure}$ and GPHH-C$_\text{Refill}$ achieved much better test performance than GPHH. This indicates that the two proposed collaborations both during route failure and refill can improve the solution quality. The effectiveness of the collaboration during refill seems to be more effective than the one during the route failure, especially for the large \emph{Uegl} instances. This indicates that the partial service of the tasks on the way to refill can greatly reduce the opportunity of route failure. Furthermore, GPHH-C obtained much better test performance than GPHH-C$_\text{RouteFailure}$ and GPHH-C$_\text{Refill}$. This demonstrates the effectiveness of combining the two collaboration activities together.

\begin{table*}[!ht]
    \footnotesize
    \centering
    \caption{The average test performance of the EDASLS, GPHH, and GPHH-C with different collaboration components (with truncated demand estimation) on the \emph{Ugdb}, \emph{Uval} and \emph{Uegl} datasets.}
    \label{tab:collres}
    \begin{tabular}{@{}ccccccc@{}}
        \toprule
         & EDASLS & GPHH & GPHH-C$_\text{RouteFailure}$ & GPHH-C$_\text{Refill}$ & GPHH-C \\
        \midrule
        \emph{Ugdb} & \textbf{277.92(2.11)} & 284.21(4.86) & 281.36(5.24) & 282.25(4.75) & 279.57(4.89) \\
        \emph{Uval} & 386.15({4.99}) & 389.73(7.57) & 387.59(7.53) & 385.52(7.28) & \textbf{383.35(6.52)} \\
        \emph{Uegl} & 13772.39({258.02}) & 13327.33(323.13) & 13227.41(351.58) & 12787.94(292.02) & \textbf{12690.39(300.82)} \\
        \bottomrule
    \end{tabular}
\end{table*}

One can see in Table \ref{tab:wdlEx} that GPHH-C performed the best overall. It significantly outperformed EDASLS on 52 out of the 81 instances, while was outperformed by EDASLS on 23, most of which are the small \emph{Ugdb} instances. GPHH-C showed a clear advantage over GPHH, GPHH-C$_\text{RouteFailure}$ and GPHH-C$_\text{Refill}$. Note that even with a single collaborative activity, GPHH-C$_\text{RouteFailure}$ and GPHH-C$_\text{Refill}$ significantly outperformed EDASLS on more instances than GPHH.

\begin{table}[!ht]
\centering
\footnotesize
\caption{The win-draw-lose results of the pairwise comparisons between the algorithms.} \label{tab:wdlEx}
\begin{tabular} {@{}c@{\hspace{3pt}}c@{\hspace{3pt}}c@{\hspace{8pt}}c@{\hspace{8pt}}c@{}}
\toprule
   & GPHH-C$_\text{RouteFailure}$ & GPHH-C$_\text{Refill}$ & GPHH & EDASLS \\ 
   \midrule
    GPHH-C & 43-38-0 & 22-58-1 & 52-28-1 & 52-6-23  \\
GPHH-C$_\text{RouteFailure}$ & --- & 7-45-29 & 22-57-2 & 41-12-28  \\
GPHH-C$_\text{Refill}$ & --- & --- & 42-38-1 & 46-7-28  \\
GPHH & --- & --- & --- & 35-15-31  \\
\bottomrule
\end{tabular}
\end{table}

\subsection{Semantic Analysis of Evolved Policies}

Eqs. (\ref{eq:rule1start})--(\ref{eq:rule1end}) show a representative policy evolved by GPHH-C for the \emph{Uegl}-s4-C instance. The policy has a promising test performance ($27494$, while the mean of GPHH-C is $28213$).
\begin{align}
    T & = T_1+T_2+\max\{T_3,T_4\}+T_5, \label{eq:rule1start} \\ 
    T_1 & = \text{CFH} / \text{FULL} - \label{eq:r1t1} \text{CR} + 0.45, \\ 
    T_2 & = - \min\{\max\{\text{CFH} / \text{FULL}, \text{SC}\}, \text{CR} - 0.45\}, \label{eq:r1t2} \\ 
    T_3 & = \text{CFH} + \min\{\text{DEM1} - \text{CFR1}, 0\}, \\
    T_4 & = \min\{\text{RQ1} + \text{SC}, \text{CFH} + \text{FRT}\}, \\
    T_5 & = (\text{FULL} - \text{FRT}) * \text{CTD}. \label{eq:rule1end}
\end{align}
We can identify the following patterns from the above policy. 
\begin{itemize}
    \item $T_1$, $T_3$ and $T_4$ have a positively correlated CFH, which implies the tasks with small CFH (i.e. close to the current location) are preferred.
    \item If there are tasks with very small CFH, then $T_2$ can be simplified to $T_2 = -\min\{\text{SC}, \text{CR}-0.45\}$. CR, which indicates the distance from the current location to the depot, and SC, which denotes the incurred cost of serving a task, remain. If the vehicle is close to the depot, then $T_2 = 0.45 - \text{CR}$, whose value is the same for all the candidate tasks. However, if the vehicle is far away from the depot, $T_2 = -\text{SC}$, and the long edged tasks are preferred.
    \item $T_3$ prefers the tasks closer to the current place. In addition, it prefers the tasks with small DEM1 and large CFR1, i.e. those nearby alternate remaining tasks with small demand whilst also distant from other prospective routes.
    \item In $T_4$, the second term prefers the tasks with small CFH. The first term may be simplified to RQ1 if the vehicle is far away from the depot (SC is cancelled out by $T_2=-\text{SC}$). In other words, if a task's closest alternative route is almost full (small RQ1), then it is less likely to be repaired by that alternative route, and thus should be served now.
    \item In $T_5$, if FULL is smaller than FRT, i.e. the route is very empty, then $\text{FULL} - \text{FRT}$ is negative, and the tasks with large CTD (far away from the depot) are preferred. On the other hand, if the route is very full, then the tasks that are closer to the depot are preferred. This is consistent with the idea of the manually designed PS5 heuristic \citep{Lacomme2004}. 
\end{itemize}

Overall, we summarise our findings from the analysis as follows.
\begin{itemize}
    \item CFH is a very important feature. The tasks with smaller CFH should be highly prioritised.
    \item The priority function should include non-linearity (e.g. min and max functions) to distinguish different situations. For example, for a relatively empty route, the tasks far away from the depot should be prioritised. For a relatively full route, the tasks close to the depot should be prioritised.
    \item GPHH-C can properly employ the global information as terminals. For example, the tasks which are less likely to be repaired by the alternative routes are prioritised in the policy.
\end{itemize}

\subsection{Illustration of Collaborative Routes}

\begin{figure*}[t]
    \centering
    \begin{minipage}{0.48\textwidth}
        \centering
        \includegraphics[width=1.0\textwidth]{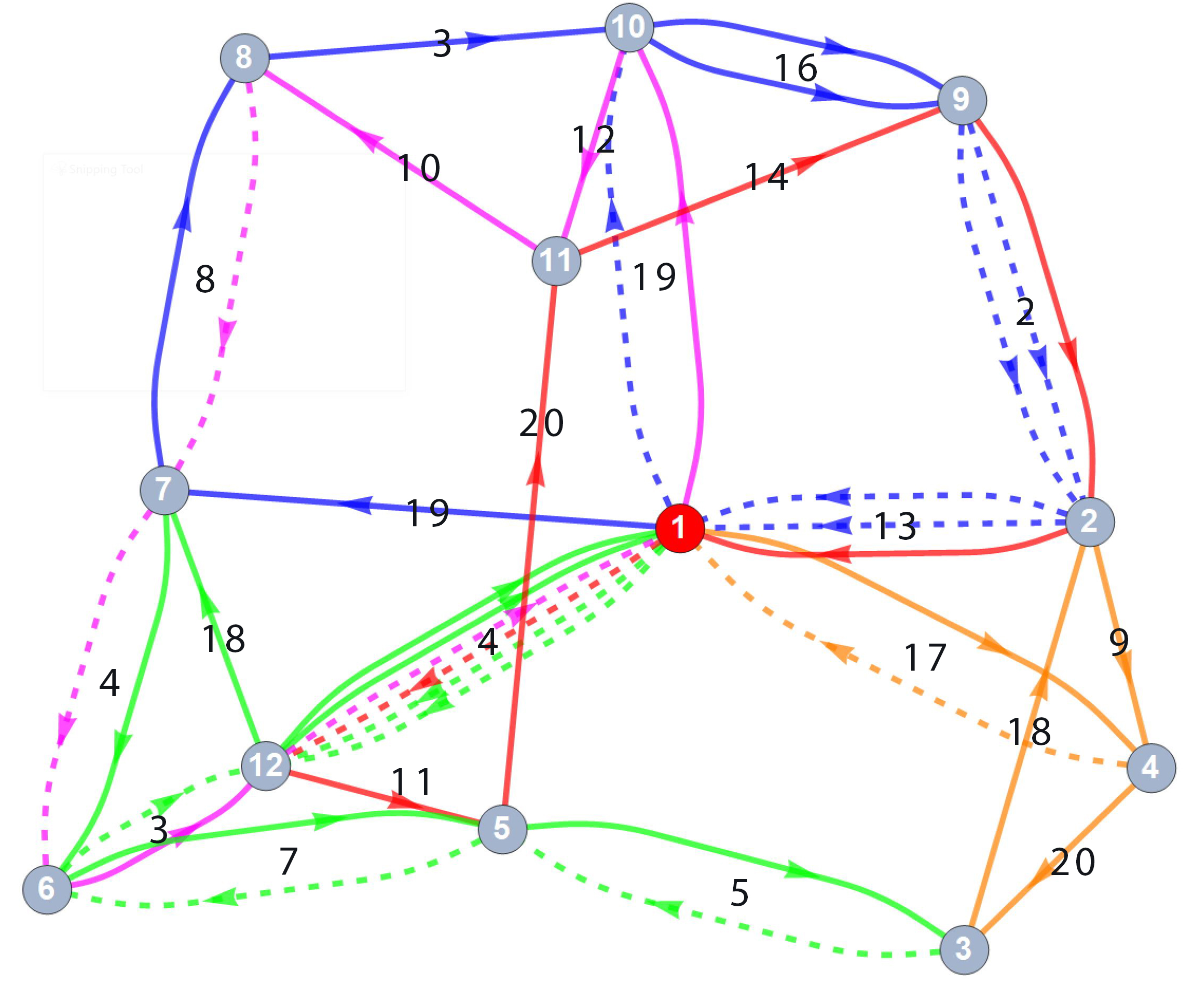} 
        \caption{A solution without collaboration generated by GPHH. Total expected cost: 393} \label{fig:GPHH_solution}
    \end{minipage}\hfill
    \begin{minipage}{0.48\textwidth}
        \centering
        \includegraphics[width=1.0\textwidth]{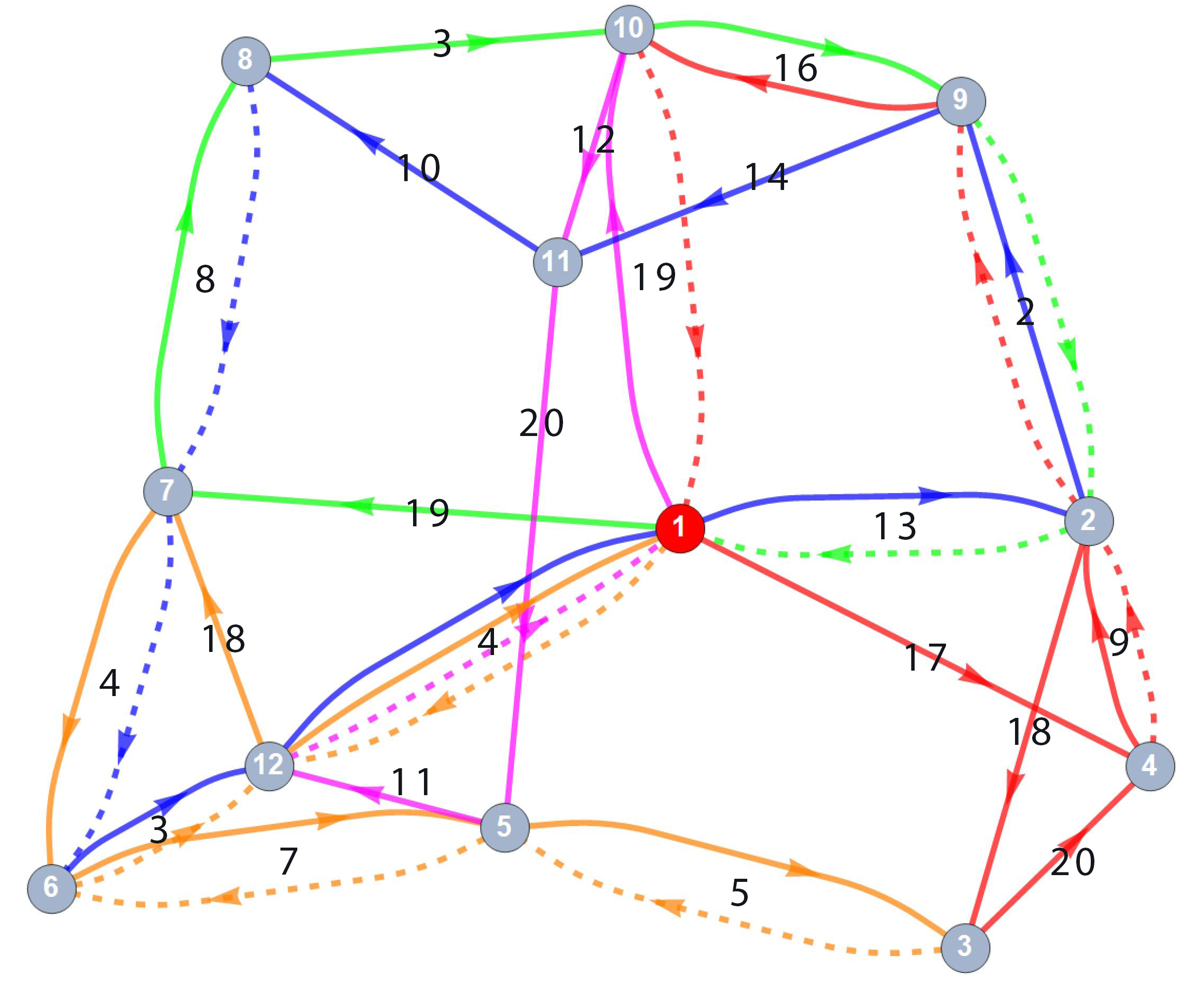} 
        \caption{A solution with collaboration generated by GPHH-C. Total expected cost: 364} \label{fig:GP-PRX_solution}
    \end{minipage}\hfill \vfill
    
    \begin{minipage}{0.48\textwidth}
        \begin{table}[H]
        \caption{The task sequence of the solution in Fig. \ref{fig:GPHH_solution}.}
        \centering
            \label{tab:GPHH_taskSeqGDB1}
            \begin{tabular} { | c | p{5cm} | } 
            \hline
            	T & \textbf{Node ID} \emph{(Served fraction)} \\ \hline
            	
                0 & \parbox[t]{0.8\textwidth}{\textbf{1} $\rightarrow$ \textbf{12} \emph{(1)} \textbf{7} \emph{(1)} \textbf{6} \emph{(1)} \textbf{5} \emph{(1)} \textbf{3} $\rightarrow$ \textbf{5} $\rightarrow$ \textbf{6} $\rightarrow$ \textbf{12} \emph{(0.943)} \textbf{1} $\rightarrow$ \textbf{12} \emph{(0.057)} \textbf{1}} \\
                
                1 & \parbox[t]{0.8\textwidth}{\textbf{1} $\rightarrow$ \textbf{12} \emph{(1)} \textbf{5} \emph{(1)} \textbf{11} \emph{(1)} \textbf{9} \emph{(1)} \textbf{2} \emph{(1)} \textbf{1}} \\
                
                2 & \parbox[t]{0.8\textwidth}{\textbf{1} \emph{(1)} \textbf{7} \emph{(1)} \textbf{8} \emph{(1)} \textbf{10} \emph{(0.869)} \textbf{9} $\rightarrow$ \textbf{2} $\rightarrow$ \textbf{1} $\rightarrow$ \textbf{10} \emph{(0.131)} \textbf{9} $\rightarrow$ \textbf{2} $\rightarrow$ \textbf{1}} \\ 
                
                3 & \parbox[t]{0.8\textwidth}{\textbf{1} \emph{(1)} \textbf{4} \emph{(1)} \textbf{3} \emph{(1)} \textbf{2} \emph{(1)} \textbf{4} $\rightarrow$ \textbf{1}} \\
                
                4 & \parbox[t]{0.8\textwidth}{\textbf{1} \emph{(1)} \textbf{10} \emph{(1)} \textbf{11} \emph{(1)} \textbf{8} $\rightarrow$ \textbf{7} $\rightarrow$ \textbf{6} \emph{(1)} \textbf{12} $\rightarrow$ \textbf{1}} \\ \hline
            \end{tabular}
\end{table}
    \end{minipage}\hfill
    \begin{minipage}{0.48\textwidth}
        \begin{table}[H]
        \caption{The task sequence of the solution in Fig. \ref{fig:GP-PRX_solution}.}
        \centering
            \label{tab:GP-PRX_taskSeqGDB1}
            \begin{tabular} { | c | p{5cm} | } 
            \hline
            	T & \textbf{Node ID} \emph{(Served fraction)} \\ \hline
            	
                0 & \textbf{1} \emph{(1)} \textbf{7} \emph{(1)} \textbf{8} \emph{(1)} \textbf{10} \emph{(0.869)} \textbf{9} $\rightarrow$ \textbf{2} $\rightarrow$ \textbf{1} \\
                
                1 & \textbf{1} \emph{(1)} \textbf{4} \emph{(1)} \textbf{2} \emph{(1)} \textbf{3} \emph{(1)} \textbf{4} $\rightarrow$ \textbf{2} $\rightarrow$ \textbf{9} \emph{(0.131)} \textbf{10} $\rightarrow$ \textbf{1} \\
                
                2 & \textbf{1} \emph{(1)} \textbf{2} \emph{(1)} \textbf{9} \emph{(1)} \textbf{11} \emph{(1)} \textbf{8} $\rightarrow$ \textbf{7} $\rightarrow$ \textbf{6} \emph{(1)} \textbf{12} \emph{(0.057)} \textbf{1} \\ 
                
                3 & \textbf{1} $\rightarrow$ \textbf{12} \emph{(1)} \textbf{7} \emph{(1)} \textbf{6} \emph{(1)} \textbf{5} \emph{(1)} \textbf{3} $\rightarrow$ \textbf{5} $\rightarrow$ \textbf{6} $\rightarrow$ \textbf{12} \emph{(0.943)} \textbf{1} \\
                
                4 & \textbf{1} \emph{(1)} \textbf{10} \emph{(1)} \textbf{11} \emph{(1)} \textbf{5} \emph{(1)} \textbf{12} $\rightarrow$ \textbf{1} \\ \hline
            \end{tabular}
\end{table}
    \end{minipage}
\end{figure*}

To understand how the vehicle collaboration improves the results, we selected a \emph{Ugdb}1 instance sample, and visualised the solutions generated by GPHH (without collaboration) and GPHH-C (with collaboration). Figs. \ref{fig:GPHH_solution} and \ref{fig:GP-PRX_solution} show the two solutions, and Tables \ref{tab:GPHH_taskSeqGDB1} and \ref{tab:GP-PRX_taskSeqGDB1} give their corresponding sequence representations. In the figures, the depot (1) is marked in red. The number associated with each edge denotes its expected traversal cost. A solid arrow indicates a service (including one with route failure), while a dashed arrow means a traversal without service.

From this example, one can see that each policy incurred the same two route failures on edges $(10,9)$ and $(12,1)$. GPHH initiates and completes these failures with the same vehicles, as shown by routes 0 and 2 of Table \ref{tab:GPHH_taskSeqGDB1}, and the same coloured routes repeating the edge in Figure \ref{fig:GPHH_solution}. GPHH-C, on the other hand, initiates the failures with one vehicle and repairs them with another, shown in Table \ref{tab:GP-PRX_taskSeqGDB1} and Figure \ref{fig:GP-PRX_solution}. For example, the $(10,9)$ edge route failure is failed by route 0 (green), then completed by route 1 (red). This collaborative effect results is a much lower expected total cost: GPHH 393 vs GPHH-C 364.

\section{Conclusions and Future Works} \label{sec:conclusion}


In this paper, we proposed a novel Solution Construction Procedure (SCP) for vehicles to solve UCARP in a completely reactive decision making process. Within this framework, we proposed two events that enabled the collaborative completion of tasks via two methods. First, when route failures occurs, the failed task may be completed by any other vehicle to allow the failing vehicle to directly service a more suitable task after refilling. Second, a vehicle can reduce the workload of another vehicle by partially serving any task on its way to refill. To reduce the rate of route failures, two demand estimation methods were proposed for use in the feasible task set filter. We then presented GPHH-C to evolve routing policies, evaluated within the SCP based framework using the proposed collaborative events. The experimental results showed that GPHH-C significantly outperformed the state-of-the-art EDASLS \citep{Wang2016} and the GPHH without Collaboration \citep{Liu2017}.

The proposed vehicle collaboration scheme is not restricted to the SCP proposed in this paper, but can be easily extended to other decision making processes such as the rollout algorithm \citep{Goodson2016} and Monte Carlo tree search \citep{sabar2015population}. Extending the collaborative scheme to other frameworks is an area of research we would like to continue with. It is not difficult to consider multi-vehicle (or multi-agent) collaboration applied to other routing problems, such as vehicle routing and scheduling problems. It may be more interesting, however, to consider its implications in other combinatorial optimisation problems, or at a higher level such as within the genetic programming process itself.

\small

\bibliographystyle{apalike}
\bibliography{Bibliography}

\end{document}